\documentclass[10pt,twocolumn,letterpaper]{article}

\usepackage{fancyhdr}
\fancypagestyle{firststyle}
{
   \fancyhf{}
   \fancyhead[C]{\footnotesize Accepted at the IEEE/CVF Conference on Computer Vision and Pattern Recognition (CVPR) 2022.}
}

\usepackage[pagenumbers]{cvpr} 

\usepackage{graphicx}
\usepackage{amsmath}
\usepackage{amssymb}
\usepackage{booktabs}
\usepackage{comment}
\usepackage{multirow}
\usepackage{import}
\usepackage{wasysym,ifsym}
\usepackage{stmaryrd}
\usepackage{float}
\usepackage[accsupp]{axessibility}  

\usepackage{bbm}
\newcommand{\Reals}{\mathbbm{R}}      

\usepackage[pagebackref,breaklinks,colorlinks]{hyperref}

\let\originalleft\left
\let\originalright\right
\def\left#1{\mathopen{}\originalleft#1}
\def\right#1{\originalright#1\mathclose{}}

\usepackage[capitalize]{cleveref}
\crefname{section}{Sec.}{Secs.}
\Crefname{section}{Section}{Sections}
\Crefname{table}{Table}{Tables}
\crefname{table}{Tab.}{Tabs.}

\def\cvprPaperID{10780} 
\def\confName{CVPR}
\def\confYear{2022}

\usepackage{xspace}
\newcommand{\name}{C-FSCIL\xspace}

 \allowdisplaybreaks 
\begin{document}
\title{Constrained Few-shot Class-incremental Learning}

{
\author{
Michael Hersche$^{1,2}$\\
{\tt\small her@zurich.ibm.com}
\and 
Geethan Karunaratne$^{1,2}$\\
{\tt\small kar@zurich.ibm.com}
\and 
Giovanni Cherubini$^{1}$\\
{\tt\small cbi@zurich.ibm.com}
\and 
Luca Benini$^{2}$\\
{\tt\small lbenini@iis.ee.ethz.com}
\and 
Abu Sebastian$^{1}$\\
{\tt\small ase@zurich.ibm.com}
\and 
Abbas Rahimi$^{1}$\\
{\tt\small abr@zurich.ibm.com}
\and
{
$^{1}$IBM Research-Zurich, $^{2}$ETH Zurich }
}
}
\maketitle
\thispagestyle{firststyle}
\begin{abstract}
Continually learning new classes from fresh data without forgetting previous knowledge of old classes is a very challenging research problem. 
Moreover, it is imperative that such learning must respect certain memory and computational constraints such as (i) training samples are limited to only a few per class, (ii) the computational cost of learning a novel class remains constant, and (iii) the memory footprint of the model grows at most linearly with the number of classes observed.
To meet the above constraints, we propose \name, which is architecturally composed of a frozen meta-learned feature extractor, a trainable fixed-size fully connected layer, and a rewritable dynamically growing memory that stores as many vectors as the number of encountered classes. 
\name provides three update modes that offer a trade-off between accuracy and compute-memory cost of learning novel classes.
\name exploits hyperdimensional embedding that allows to continually express many more classes than the fixed dimensions in the vector space, with minimal interference.
The quality of class vector representations is further improved by aligning them quasi-orthogonally to each other by means of novel loss functions.  
Experiments on the CIFAR100, miniImageNet, and Omniglot datasets show that \name outperforms the baselines with remarkable accuracy and compression.
It also scales up to the largest problem size ever tried in this few-shot setting by learning 423 novel classes on top of 1200 base classes with less than 1.6\% accuracy drop. Our
code is available at \url{https://github.com/IBM/constrained-FSCIL}.

\end{abstract}

\section{Introduction}
Deep convolutional neural networks (CNNs) have achieved remarkable success in various computer vision tasks, such as image classification~\cite{AlexNet_NIPS2012,DeeperCNN_CVPR2015,ResNet_CVPR2016,BiT_ECCV2020}, stemming from the availability of large curated datasets as well as huge computational and memory resources.
This, however, poses significant challenges for their applicability to smart agents deployed in new and dynamic environments, where there is a need to continually learn about novel classes from very few training samples, and under resource constraints.  
We consider the challenging scenario of learning from an online stream of data, including never-seen-before-classes, where we impose constraints on the sample size, computational cost, and memory size.

\begin{figure}
    \centering
    \def\svgwidth{\linewidth}
    \fontsize{7}{9}
    \selectfont
    \includegraphics[width=\linewidth]{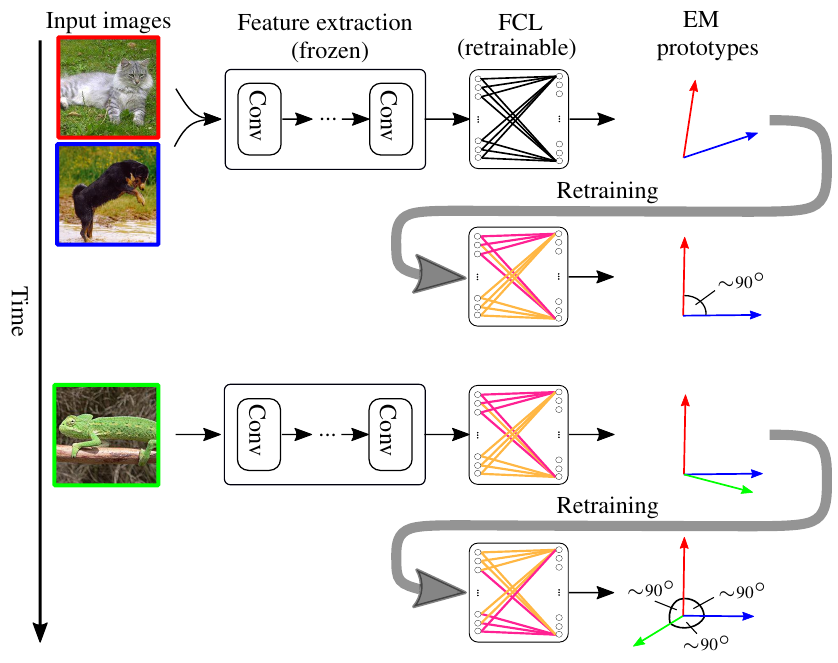}
    \caption{
    Overview of \name which maps input images to quasi-orthogonal prototypes such that the prototypes of different classes encounter small interference.
    }
  \label{fig:figure1}
\end{figure}

Let us first focus on the sampling constraint of training data. 
Inspired by human-like sequential learning, classical connectionist networks can be naively trained sequentially, e.g., first on a set of old classes, and then on a set of novel classes, whereby the training dataset of old classes is no longer available. 
As a result, the new learning may interfere catastrophically with the old learning by overwriting weights involved in representing the old learning (and thereby forgetting)~\cite{catastrophicForgetting_1989}.
This effect is known as \emph{catastrophic interference}, or \emph{catastrophic forgetting}, and causes the classification accuracy to deteriorate~\cite{catastrophicForgetting_1989,catastrophicForgetting_ICLR14}.
To address the catastrophic forgetting problem, research efforts have been directed to, e.g., freezing parts of the network weights, while simultaneously growing other parts of the network to extend the learning ability.
Among them, class-incremental learning (CIL)~\cite{iCaRL_CVPR2017,E2E_CIL_ECCV2018,Rebalancing_CIL_CVPR2019,LargeScl_CIL_CVPR2019,LatenReplay_IROS2020} aims to learn a unified classifier in which the encountered novel classes---that were not seen before in the continual data stream---are added into the recognition tasks without forgetting the previously observed classes.
One step further, very recently, few-shot CIL (FSCIL)~\cite{FSCIL_CVPR2020,VecQ_FSCIL_ICLR2021,FlatMin_FSCIL_NIPS2021,Evolv_FSCIL_CVPR2021,Synthesized_FSCIL_ICCV2021,Semantic_FSCIL_CVPR2021,AttAttrac_FSCIL_NIPS19,ecai2020} algorithms have been proposed to continually extend learning to novel classes with only a few data samples. 
Requiring FSCIL to be trained with very few novel training samples makes it more challenging compared to CIL, which usually learns new classes from large-scale training samples.

To facilitate few-shot learning, memory-augmented neural networks (MANNs) \emph{separate} information processing from memory storage~\cite{graves14,weston15,sukhbaatar15,graves16,MANN_pmlr2016,KarunaratneHDAugmented2021,GenKV_TNNLS22}. 
MANNs incorporate an \emph{explicit} memory (EM) into an embedding network such that the network can write the embedding of few data samples to the memory as class prototypes, and read from these individual entries during inference. 
This separation enables the network to offload new prototypes to the EM, where they do not endanger the previously learned prototypes to be overwritten, leading to remarkable accuracy with few classes (typically 20 classes).
However, this neat feature of MANNs has not been exploited in CIL or FSCIL. 
This is mainly due to the fact that the interference between different class prototypes increases with a growing number of classes, as experienced in FSCIL.
To allow continual learning in MANNs, the representational power of the embedding network should be naturally extensible to a large number of classes with minimal interference. 
This critical requirement can be met by exploiting hyperdimensional computing~\cite{KanervaHyperdimensional2009,VSA03,Plate_TNN95}, where classes can be represented by random high-dimensional vectors with a dimensionality in the order of thousands.
This new combination of MANNs and hyperdimensional computing has been recently developed in~\cite{KarunaratneHDAugmented2021} for few-shot learning.

Hyperdimensional computing is characterized by the following properties: 
(i) A randomly chosen vector is quasi-orthogonal to other random vectors with very high probability (the ``curse'' of dimensionality), therefore the representation of a novel class is not only incremental to the old learning but also causes minimal interference.
This phenomenon is known as concentration of measure~\cite{LedouxConcentration2001}, with the peculiar property that pseudo-orthogonality converges to exact orthogonality with increasing dimensionality. 
(ii) The number of such quasi-orthogonal vectors grows exponentially with the dimensionality, which provides a sufficiently large capacity to accommodate novel classes over time.
(iii) Counterintuitively, quasi-orthogonal vectors can still encode semantic information.
We can describe a concept in a scene (e.g., a black dog) with a vector by binding quasi-orthogonal atomic vectors ($\mathbf{x}_{\mathrm{black}}\varoast \mathbf{x}_{\mathrm{dog}}$), which is quasi-orthogonal to all other involved vectors (atomic and composite). 
The bound vector can be decomposed to $\mathbf{x}_{\mathrm{black}}$ and $\mathbf{x}_{\mathrm{dog}}$, revealing the semantic relation between a black dog and a black cat (both include $\mathbf{x}_{\mathrm{black}}$)~\cite{frady2020resonator,hersche2022nvsa}.
In fact, fixed quasi-orthogonal vectors have been successfully used as class vectors achieving improved performance in the supervised classification tasks~\cite{FixClassifier_ICLR18,hersche2022nvsa}.

In this paper, we enhance MANNs by exploiting the representational power of hyperdimensional computing to perform FSCIL.
During FSCIL operations, it is constrained to either no gradient updates or a small constant number of iterations for learning novel classes, and a linear growth in the memory size with respect to the encountered classes. 
The contributions of this work are summarized as follows:

First, we propose \name, which is architecturally composed of a frozen feature extractor, a trainable fixed-size fully connected layer, and a rewritable dynamically growing EM that stores as many vectors as the number of classes encountered so far (See Fig.~\ref{fig:figure1}).
The frozen part is separated from the growing part by inserting the fully connected layer, which outputs class vectors in a hyperdimensional embedding space whose dimensionality remains fixed, and is therefore independent of the number of classes in the past and future.
The feature extractor is a CNN that is meta-learned by proper sharpened attention, which strives to represent dissimilar images with quasi-orthogonal vectors.   
The \name architecture is presented in Section~\ref{sec:arch}.

Second, \name with three update modes offers a trade-off between accuracy and the compute-memory cost of learning a novel class. 
The simple yet powerful Mode\,1 creates and updates an averaged prototype vector as the mean-of-exemplars in the EM, without any gradient computation (see Section~\ref{sec:avg_prototypes}).
Mode\,2 bipolarizes the prototype vectors, and retrains the fully connected layer without exceeding a small constant number of iterations; for retraining it requires storing an averaged activation pattern for every class in a globally averaged activation (GAA) memory (see Section~\ref{sec:avg_prototypes_retraining}).
Mode\,3 nudges the averaged prototype vectors to align them quasi-orthogonally to each other, while remaining in close proximity to the original averaged prototypes, using a combination of novel loss functions. 
It then retrains the fully connected layer without exceeding the constant number of iterations (see Section~\ref{sec:nudged_prototypes}).  
%

Third, \name leads to higher accuracy, compute-memory efficiency, and scalability compared to FSCIL baselines, as shown in Section~\ref{sec:expr}.
Experiments on the CIFAR100, miniImageNet, and Omniglot datasets demonstrate that \name outperforms the baselines even in Mode\,1 where prototypes are simply obtained in one pass without any gradient-based parameter updates.
In Omniglot, \name scales up to 1623 classes, whereby 423 novel classes are incrementally added to 1200 base classes, with less than 2.6\%, 1.4\%, and 1.6\% accuracy drops when using Modes\,1,\,2, and\,3, respectively. 
Thanks to the quasi-orthogonality of the class prototypes in the EM, they can be compressed by $2\times$, causing 1.7\%--3.5\% accuracy drop during the course of FSCIL.

\section{Related Work}
\subsection{MANNs}
MANNs have been used in one- and few-shot learning tasks ~\cite{weston15,MANN_pmlr2016,matching_net,KarunaratneHDAugmented2021,GenKV_TNNLS22}.
First, their embedding network is meta-learned on a set of initial categories of so-called base classes.
After meta-learning, the network is ready to provide new representations for never-seen-before categories (i.e., novel classes) that can be independently stored, or retrieved via the EM. %
In these works, training is done on the base classes and inference is done solely on the novel classes, as opposed to the union of the two sets of classes as required by FSCIL. 
Further, the size of the EM is set by $c$ vectors~\cite{MANN_pmlr2016,matching_net}, or $c \times k$ vectors~\cite{KarunaratneHDAugmented2021,Wu_2018_CVPR,Wu_2018_ECCV} where $c$ and $k$ represent the number of classes and training examples per class, respectively. 
The contents of the EM can be compressed using outer products with randomized labels~\cite{GenKV_TNNLS22}. 
In a similar vein, \name superposes two class vectors to store $c/2$ vectors in the EM, as described in the following.

\subsection{Rehearsal and Pattern Replay}
To address the catastrophic forgetting issue in CIL, rehearsal methods store some old training data, and replay them while learning novel classes during the incremental stage~\cite{GradientEM_NIPS17,iCaRL_CVPR2017, Bang2021rainbow,prabhu2020gdumb,borsos2020coresets, Liu_2020_CVPR}.
For instance, a flexible number of exemplars per class are stored in an \emph{episodic memory}.
It provides the training data of the old classes as well as the currently available classes to a classifier to incrementally learn novel classes.
The memory footprint of the episodic memory is kept bounded by relying on entry/exit criteria~\cite{iCaRL_CVPR2017,Bang2021rainbow,prabhu2020gdumb,borsos2020coresets, Liu_2020_CVPR}. 
Alternatively, the exemplars of past classes can be generated using GANs with extra computational cost~\cite{cong2020GANforgetting}.
The storage and computational overhead of the rehearsal methods can be reduced by \emph{latent replay}~\cite{LatenReplay_IROS2020} and its TinyML evolution~\cite{PULP_LR_2021}.
Instead of storing a portion of past exemplars in the input space, the latent replay method stores the corresponding activation patterns from an intermediate layer~\cite{LatenReplay_IROS2020}. 
The stored activations are replayed to retrain all the layers above the latent replay layer. 

Our \name avoids the rehearsal and pattern replay overheads because it does not need any access to the previous data in any form.
Instead, it maintains a highly compressed, minimal amount of past knowledge, either in the EM (Mode\,1) or in the GAA memory (Modes\,2-3), which is similar or smaller than the compared methods.
Note that in Modes\,2-3 only storing the GAA memory is sufficient as it contains $c$ compressed activations, from which the EM can be \emph{rematerialized} on the fly.   
For discriminating $c$ classes, one needs to maintain at least $c$ prototypes (class vectors); thus, a linear increase with the number of classes is unavoidable. 
In addition, if the memory constraints become even tighter, our quasi-orthogonal design comes to rescue to further compress either of the memories by binding each vector with a randomly drawn key and superimposing two key-prototype pairs~\cite{Plate_TNN95}, yielding twofold compression with moderate accuracy drop (See Appendix~\ref{appendix:compression}).

\subsection{Class Imbalance and Reparameterizations}
Another issue in CIL is the class imbalance problem, where the norm or the bias can be unbalanced for the classes observed later in the incremental stage, causing the network’s prediction to be biased towards novel classes~\cite{LargeScl_CIL_CVPR2019,Rebalancing_CIL_CVPR2019}.
Learning novel classes might also interfere with the past classes.
To mitigate the impact of class imbalance, recent CIL approaches adopt the cosine distance metric~\cite{Rebalancing_CIL_CVPR2019}, or add a bias correction layer~\cite{LargeScl_CIL_CVPR2019} to avoid the norm and bias imbalance.
In a similar vein, a weight aligning method has been proposed to align the norms of the weight vectors for novel classes to those for past classes~\cite{WeightAligning_CVPR2020}.
Another source of perturbation is the gradient from novel class observations that can affect past classes to change their weights.
Regularization strategies have been aimed at avoiding such forgetting~\cite{Regular_CatastForget_2017}, but they have been recently shown to be insufficient in CIL~\cite{Regular_CIL_2021}.
Various maskings have been proposed to apply the gradient only to a subset of classes during back-propagation~\cite{LastLayer_CIL_2021}.

We avoid these issues systematically in \name, where hyperdimensional quasi-orthogonal vectors are assigned to each and every class with the aim of reducing interference.
This cannot be achieved with other methods that replace the fully connected layer with a fixed Hadamard~\cite{FixClassifier_ICLR18} or identity~\cite{IdentityMatx_2020} matrix, because they fail to support a larger number of classes than the vector dimensionality of the layer they are connected to.
Our prototypes are stored in the EM, the cosine similarity is used for their comparison, and they can be selectively updated.
The alignment to prototypes can be improved by proper retraining of the fully connected layer whose structure remains fixed and independent of the number of classes (See Fig.~\ref{fig:figure1}).

\subsection{Few-shot Class-incremental Learning (FSCIL)}

Very recently, FSCIL~\cite{FSCIL_CVPR2020,VecQ_FSCIL_ICLR2021,FlatMin_FSCIL_NIPS2021,Evolv_FSCIL_CVPR2021,Synthesized_FSCIL_ICCV2021,Semantic_FSCIL_CVPR2021} has been proposed for tackling CIL with very few training samples through a number of incremental sessions.
Various solutions have been proposed, such as exploiting a neural gas network~\cite{FSCIL_CVPR2020}, a graph attention network~\cite{Evolv_FSCIL_CVPR2021}, semantic word embeddings~\cite{Semantic_FSCIL_CVPR2021}, and vector
quantization~\cite{VecQ_FSCIL_ICLR2021}.
Specifically, a pretrained backbone is decoupled from a non-parametric class mean classifier whose weights can be progressively adapted across sessions by a graph attention network~\cite{Evolv_FSCIL_CVPR2021}.
\name, even with the simple Mode\,1, which does not need retraining, sets a new state-of-the-art accuracy compared to the previous FSCIL~\cite{FSCIL_CVPR2020,VecQ_FSCIL_ICLR2021,FlatMin_FSCIL_NIPS2021,Synthesized_FSCIL_ICCV2021,Semantic_FSCIL_CVPR2021,Evolv_FSCIL_CVPR2021}, and demonstrates its extensible representation by supporting the maximum number of encountered classes (1623) in this setting. 
In fact, all previous works have used up to 200 classes.
To be able to do retraining in the other modes, \name requires the GAA memory to store as many compressed activation patterns as there are encountered classes.
It allows to stably retrain with any combination of sessions or classes over time, and with an improved accuracy compared to Mode\,1.

\section{Notations and Preliminaries}
\subsection{Problem Formulation}
FSCIL sequentially provides training sets $\mathcal{D}^{(1)}$, $\mathcal{D}^{(2)}$, ..., $\mathcal{D}^{(s)}$,...,$\mathcal{D}^{(S)}$, where $\mathcal{D}^{(s)}:= \lbrace(\mathbf{x}^{(s)}_n, y^{(s)}_n)\rbrace _{n=1}^{|\mathcal{D}^{(s)}|}$ with input data $\mathbf{x}^{(s)}_n$, e.g., an image, and corresponding ground-truth labels $y^{(s)}_n$. 
The labels $y^{(s)}_n\in\mathcal{C}^{(s)}$ are mutually exclusive across different training sets, i.e., $\forall i\neq j, \mathcal{C}^{(i)}\cap \mathcal{C}^{(j)}=\emptyset$. 
The training sets are denoted as \emph{sessions}, whereby the first training set is denoted as the \emph{base session}, providing a larger number of training examples and classes. 
The goal of the base session is to learn a meaningful representation, where we can distinguish between the ways (i.e., classes) in $\mathcal{C}^{(1)}$. 
The following sessions provide training sets $\mathcal{D}^{(s)}$ of size $|\mathcal{D}^{(s)}|=c\cdot k$, where $c=|\mathcal{C}^{(s)}|$ is the number of ways and $k$ the number of training samples per way, hence we call it $c$-way $k$-shot. 
In a given session $s$, one only has access to the corresponding training set $\mathcal{D}^{(s)}$; the training sets of the previous sessions $1,2,...,s-1$ are not available anymore. 
The model is tested on a session-specific evaluation set $\mathcal{E}^{(s)}$ that contains samples from all previous sessions as well as the current one, i.e., $\forall j<i, \mathcal{E}^{(j)} \subset \mathcal{E}^{(i)}$. 
We denote the set of classes that have to be covered during session $s$ as $\tilde{\mathcal{C}}^{(s)}:=\cup_{i=1}^s \mathcal{C}^{(i)}$.

\section{Proposed Method: \name}
\subsection{Architecture}
\label{sec:arch}
\name consists of three main components: a feature extractor, a fully connected layer, and an EM.
It can also have an auxiliary GAA memory, which will be described in Section~\ref{sec:avg_prototypes}. 
The feature extractor maps the samples from the input domain $\mathcal{X}$ to a feature space: $f_{\mathbf{\theta}_1} : \mathcal{X} \rightarrow \Reals^{d_f}$, where $\mathbf{\theta}_1$ are the feature extractor's learnable parameters.
As the feature extractor, we use a five-layer CNN, or a ResNet-12, to map an input image to the feature space. 
To form the embedding network with a hyperdimensional distributed representation, the feature extractor is connected to a fully connected layer $g_{\mathbf{\theta}_2}: \Reals^{d_f} \rightarrow \Reals^d$, containing $\theta_2 \in \Reals^{d \times d_f}$ trainable parameters where $d \leq 512$.
Note that $d$ should be large enough to ensure that the expected similarity between randomly drawn $d$-dimensional vectors is approximately zero with a very high probability~\cite{KanervaHyperdimensional2009}, but preferably $d < |\tilde{\mathcal{C}}^{(S)}|$.   
We denote the union of the trainable parameters in the feature extractor and the fully connected layer as $\mathbf{\theta}=(\mathbf{\theta}_1, \mathbf{\theta}_2)$. 

\begin{figure}
    \centering
    \def\svgwidth{\linewidth}
    \fontsize{7}{9}
    \selectfont
    \includegraphics[width=\linewidth]{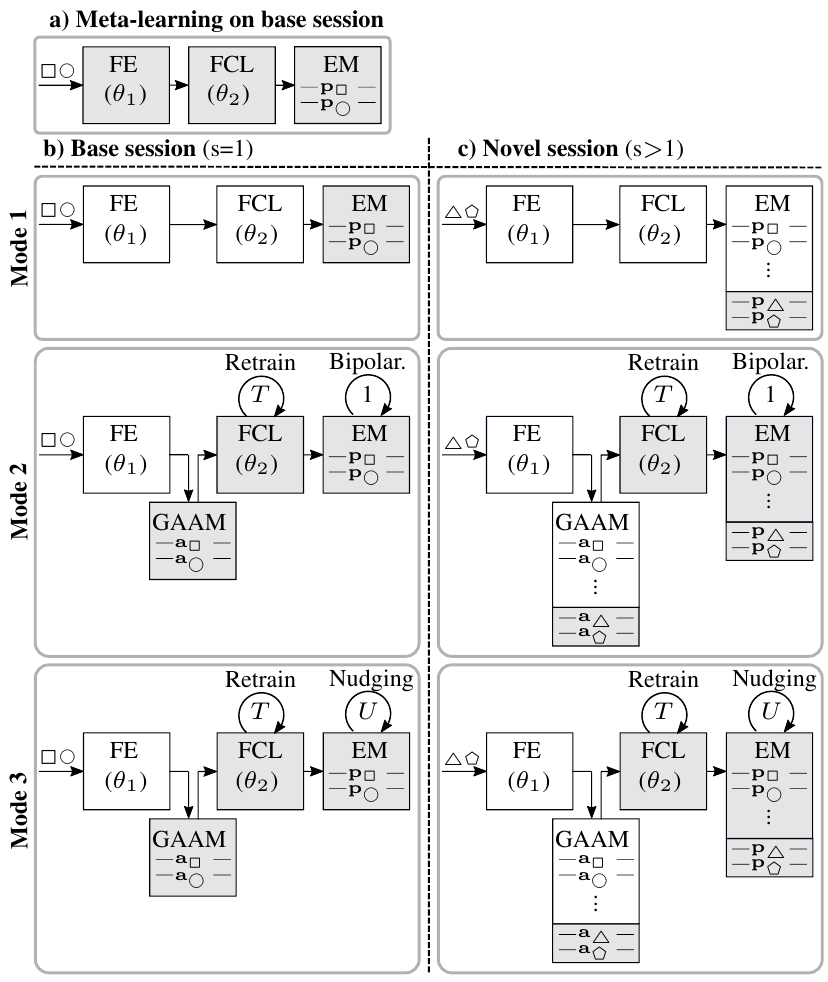}
    \caption{The \name architecture consists of a feature extraction (FE), a fully connected layer (FCL), and an explicit memory (EM).
    It also requires a globally average activation memory (GAAM) only for Modes\,2 and 3.
    It is shown how these components will evolve during \textbf{a)} meta-learning, \textbf{b)} base session, and \textbf{c)} novel sessions; the updated components are marked in gray while the frozen ones are kept in white.   
    The three update modes are shown as well, with emphasis on which components they update and for how many iterations ($1$, $T$, and $U$), and which memory they expand linearly with the number of classes.
    All three Modes require meta-learning prior to their base session.
    }
    \label{fig:concept_figure}
\end{figure}

The fully connected layer produces a support vector for every training input.
These support vectors are combined to compute a set of $d$-dimensional prototype vectors to be stored in the EM (more details in Section~\ref{sec:modes}).
Besides storing the prototype vectors, the EM contains a value memory of one-hot labels, and it is overall referred to as a key-value memory.
The prototype vectors stored in the EM are not accessed by stating a discrete address, but by comparing the cosine similarities between a query from the embedding network and all the prototype vectors. 
Given a query $\mathbf{x} \in \mathcal{E}^{(s)}$ and prototypes $\mathbf{P}^{(s)}:=(\mathbf{p}_1, \mathbf{p}_2, ..., \mathbf{p}_{|\tilde{\mathcal{C}}^{(s)}|})$, we compute the score $l_i$ for class $i\in \tilde{\mathcal{C}}^{(s)}$ as follows: 
\begin{align}
    l_i = \mathrm{cos}\left(\mathrm{tanh} \left(g_{\mathbf{\theta}_2}\left(f_{\mathbf{\theta}_1}\left(\mathbf{x}\right)\right) \right), \left(\mathrm{tanh} \left(\mathbf{p}_i\right)\right)\right), \label{eq:score}
\end{align}
where $\tanh(\cdot)$ is the hyperbolic tangent function and $\mathrm{cos}(\cdot,\cdot)$ the cosine similarity. 
The hyperbolic tangent has proven to be a useful non-linearity in the hyperdimensional MANNs~\cite{KarunaratneHDAugmented2021}, limiting the norm of activated prototypes and embedding outputs. 
Moreover, the cosine similarity resolves the norm and bias issues usually encountered in FSCIL by focusing on the angle between the activated prototypes and embedding outputs, while ignoring their norm~\cite{Rebalancing_CIL_CVPR2019,LastLayer_CIL_2021}. 
Overall, this forms a content-based attention mechanism between the embedding network and the EM by computing a similarity score for each memory entry with respect to a given query.
This attention vector is sharpened by a soft absolute sharpening function $\epsilon(\cdot)$ leading to quasi-orthogonality~\cite{KarunaratneHDAugmented2021}.
See Appendix~\ref{sec:loss}.
The resulting attention vector serves to read out the value memory.

The embedding network $(\mathbf{\theta}_1, \mathbf{\theta}_2)$ and the EM are meta-learned by solving various few-shot problem sets that gradually enhance the quality of the mapping (see Fig.~\ref{fig:concept_figure}a). 
This is done by gradually learning to assign nearly quasi-orthogonal vectors to different image classes, and mapping them far away from each other in the hyperdimensional space.
This space does not run out of such dissimilar vectors for newly encountered classes.
To improve inter-class separation, \name builds on top of the meta-learning setup and offers three modes for updating that are described in the following section.

\subsection{Update Modes}
\label{sec:modes}
\name does not need to keep track of all class representations in a centralized fashion. 
Rather, it relies on the hyperdimensional embedding space to incrementally create maximally separable classes with three different techniques that trade off memory-compute cost for improved accuracy.
In the first mode, the embedding network is already enforced (during meta-learning) to provide a valid quasi-orthogonal vector for a novel class, as described in Section~\ref{sec:avg_prototypes}.
This mode learns online without any gradient-based parameter update.
In the second mode, the averaged prototypes are bipolarized to achieve better quasi-orthogonality, followed by the retraining of the fully connected layer (Section~\ref{sec:avg_prototypes_retraining}). 
In the third mode, the quality of separation is further improved by prototype nudging and fine-tuning of the fully connected layer (Section~\ref{sec:nudged_prototypes}).
The three modes are illustrated in Fig.~\ref{fig:concept_figure}.
None of them update the weights in the feature extractor after the meta-learning.

\subsubsection{Mode 1: Averaged Prototypes}
\label{sec:avg_prototypes}
Earlier works on few-shot learning networks~\cite{matching_net,snell17} have proposed the use of class means as a single prototype per class, and then classified a query to the nearest prototype.
Inspired by these networks, we also represent each class by a single prototype vector by computing the class-wise average over all the support vectors that belong to a class. 
The prototypes are stored in the EM, which continuously expands over time as new prototypes are computed in the new sessions (see Fig.~\ref{fig:concept_figure}b and c; Mode\,1). 
The EM stores prototypes $\mathbf{P}^{(s)}=(\mathbf{p}_1, \mathbf{p}_2, ..., \mathbf{p}_{|\tilde{\mathcal{C}}^{(s)}|}), \mathbf{P}^{(s)}\in \Reals^{d\times|\tilde{\mathcal{C}}^{(s)}| }$ of all classes that have been exposed to the model so far, whereby a prototype for class $i$ is determined as follows:
\begin{align}
    \mathbf{p}_i(\mathcal{D}^{(s)},\theta) &=\frac{1}{k} \sum_{\substack{n = 1 \\  \mathrm{s.t.}\, y_n^{(s)}=i}} ^{|\mathcal{D}^{(s)}|} g_{\mathbf{\theta}_2}\left(f_{\mathbf{\theta}_1}\left(\mathbf{x}^{(s)}_n\right)\right)\label{eq:proto1}\\
    &=g_{\mathbf{\theta}_2}\left(\frac{1}{k} \sum_{\substack{n = 1 \\  \mathrm{s.t.}\, y^{(s)}_n=i}} ^{|\mathcal{D}^{(s)}|} f_{\mathbf{\theta}_1}\left(\mathbf{x}^{(s)}_n\right)\right)\label{eq:proto2}\\
    &=g_{\mathbf{\theta}_2}\left(\mathbf{a}_i(\mathcal{D}^{(s)},\mathbf{\theta}_1)\right), \forall i \in \mathcal{C}^{(s)}. 
\end{align}
The linearity of $g_{\mathbf{\theta}_2}(\cdot)$ enables the step from \eqref{eq:proto1} to \eqref{eq:proto2}. 
Each $\mathbf{a}_i$, as a $d_f$-dimensional compressed vector, represents the globally averaged activations of class $i$, and allows the determination of the corresponding prototype using $g_{\mathbf{\theta}_2}(\cdot)$.
The globally averaged activation (GAA) memory keeps track of all past averaged activations $\mathbf{A}^{(s)}:=\left(\mathbf{a}_1,\mathbf{a}_2,...,\mathbf{\mathbf{a}_{|\tilde{\mathcal{C}}^{(s)}|}}\right),\mathbf{A}^{(s)} \in \Reals^{d_f \times |\tilde{\mathcal{C}}^{(s)}|} $. 
Mode\,1 does not need the GAA memory, because the GAA memory maintains the compressed activations of the past classes that will be needed for retraining in the next two modes.

\subsubsection{Mode 2: Retraining on Bipolarized Prototypes}
\label{sec:avg_prototypes_retraining}

As the number of prototypes increases with the new sessions, they fall short in discriminating between different classes due to insufficient inter-class separability. 
To this end, we adjust the prototypes and the fully connected layer to better guide the activations that are provided by the frozen feature extractor. 
We follow a two-stage strategy, where we first adjust the prototypes and then retrain the fully connected layer to align the averaged activations with the newly adjusted prototypes.

The first step tries to create separation between nearby prototype pairs, which optimally yields close to zero cross-correlation between the prototypes pairs. 
A computationally cheap yet effective option is to add some sort of noise to the prototypes, e.g., quantization noise. 
We propose to quantize the prototypes to bipolar vectors by applying the element-wise sign operation, defined as $\mathbf{K}^* = \mathrm{sign}(\mathbf{P}^{(s)})$.

Next, the embedding has to be retrained such that its output aligns with the bipolarized prototypes. 
Instead of attempting to optimize every training sample, we aim to align the globally averaged activations available in the GAA memory, defined in~\eqref{eq:proto1}, with the bipolarized prototypes ($\mathbf{K}^{*}$). 
The final fully connected layer of the embedding network has the task of mapping localist features from the feature extractor to a distributed representation.
Hence, we find that exclusively updating the parameters of the fully connected layer $\mathbf{\theta}_{2}$ is sufficient, while the parameters of the feature extraction $\mathbf{\theta}_1$ are kept frozen during retraining.
Due to the averaged prototype-based retraining and the linearity of the fully connected layer, it is sufficient to pass the averaged activations from the GAA memory through the fully connected layer.

The fine-tuning of the fully connected layer is based on iterative updates using a loss function that strives to maximize the similarity between the averaged activations and the bipolar prototypes: 
\begin{gather}
\mathcal{L}_\text{F}\left(\mathbf{\theta}_2^{(t)},\mathbf{K}^*,\mathbf{A}\right) =-\sum_{i=1}^{|\tilde{\mathcal{C}}^{(s)}|} \mathrm{cos}\left( \mathrm{tanh}(\mathbf{k}_{i}^* ),\mathrm{tanh}(g_{\theta_2^{(t)}}(\mathbf{a}_i))\right)\label{eq:backprop_loss}\\
\mathbf{\theta}_{2}^{(t+1)} = \mathbf{\theta}_2^{(t)} - \beta \frac{\partial \mathcal{L}_\text{F}\left(\mathbf{\theta}_2^{(t)},\mathbf{K}^*,\mathbf{A}\right)}{\partial \mathbf{\theta}_{2}^{(t)}}\label{eq:retrain_update}
\end{gather}
where $\beta$ is the update rate. 

After $T$ iterations of parameter updates, the final prototypes $\mathbf{P^{*}}$ are determined by passing the globally averaged activations through the fully connected layer one last time.
The fully connected layer was already fine-tuned on the quasi-orthogonal bipolarized prototypes and therefore these final generated prototypes also tend to be quasi-orthogonal. 
Moreover, the final prototypes provide a better alignment than the earlier bipolarized prototypes.
Fig.~\ref{fig:concept_figure} illustrates this process in Mode\,2.

\subsubsection{Mode 3: Retraining on Nudged Prototypes}
\label{sec:nudged_prototypes}
This mode proposes an improved prototype alignment strategy based on solving an optimization problem instead of simply bipolarizing the prototypes. 
Given a set of initial prototypes, generated by the meta-learned embedding network, \name first optimizes them to yield the nudged prototypes.
\name then fine-tunes the fully connected layer to align it with the nudged prototypes. 

We aim to find the nudged prototypes that simultaneously
a) improve the inter-class separability by attaining a lower similarity between the pairs of nudged prototype vectors, and b) remain close to the initial averaged prototypes generated by the meta-learned embedding network. 
%
We set the initial nudged prototypes to the current prototypes stored in the EM, i.e., $\mathbf{K}^{(0)}:=\mathbf{P}^{(s)}$.  
The nudged prototypes are then updated $U$ times in a training loop to find an optimal set of prototypes unique to the given $\mathbf{A}^{(s)}$ available in the GAA memory (see Fig.~\ref{fig:concept_figure} Mode\,3).

The updates to the prototypes are based on two distinct loss functions that aim to meet the two aforementioned objectives. 
The first main objective is to decrease the inter-class similarity, which is achieved by minimizing the cross-correlation between the prototypes: 
\begin{align}
\mathcal{L}_\text{O}(\mathbf{K}^{(u)}) &= \sum_{\substack{i,j = 1 \\ i\neq j}}^{|\tilde{\mathcal{C}}^{(s)}|} \sigma\left(\mathrm{cos}\left(\mathrm{tanh}(\mathbf{k}_i^{(u)}),\mathrm{tanh}(  \mathbf{k}_j^{(u))})\right)\right),  \label{eq:offdiag_corr}
\end{align}
The activation function $\sigma(\cdot)$ penalizes prototype pairs with large absolute cross-correlations: 
\begin{align}
    \sigma(c) := e^{\alpha \cdot c}+ e^{- \alpha \cdot c} -2, \forall c\in \Reals,\label{eq:doubleexp}
\end{align}
where $\alpha =4$ controls the steepness of the loss. 
Optimally, all of the prototypes would be orthogonal due to having zero cross-correlation, which is not possible when the growing number of classes exceeds the number of dimensions.
Instead, we focus on finding the quasi-orthogonal vectors that can meet the constraint of $d<|\tilde{\mathcal{C}}^{(s)}|$.
There is a large number of quasi-orthogonal vectors that can be found by minimizing the proposed loss function $\mathcal{L}_\text{O}$.

The second objective is to keep the updated prototypes similar to the initial prototypes $\mathbf{K}^{(0)}$. 
This avoids significant deviations from the original representations of the initial base categories on which the embedding network was trained during meta-learning.
We retain high similarity between the currently updated and the initial prototypes by adding the following loss function:  
\begin{align}
\mathcal{L}_\text{M}(\mathbf{K}^{(u)},\mathbf{K}^{0}) &= -\sum_{i=1}^{|\tilde{\mathcal{C}}^{(s)}|} \mathrm{cos}\left(\mathrm{tanh}(\mathbf{k}_i^{(u)}),\mathrm{tanh}(  \mathbf{k}_i^{(0))})\right).  \label{eq:meta_deviation}
\end{align}

Finally, the nudged prototypes are updated for $U$ iterations, whereby one update is defined as:
\begin{align}
\mathbf{K}^{(u+1)} &= \mathbf{K}^{(u)} - \gamma \frac{\partial ({{\mathcal{L}}_\text{O}(\mathbf{K}^{(u)})+{\mathcal{L}}_\text{M}}(\mathbf{K}^{(u)},\mathbf{K}^{(0)}))}{\partial \mathbf{K}^{(u)}}, \label{eq:prec_loop_update}
\end{align}
where $\gamma$ denotes the update rate. 
The final nudged prototypes $\mathbf{K}^*:=\mathbf{K}^{(U)}$ will be used to retrain the fully connected layer for $T$ iterations using the loss~\eqref{eq:backprop_loss} and the update rule~\eqref{eq:retrain_update}.
Akin to Mode\,2, the final prototypes are determined by passing the globally averaged activations through the fully connected layer after retraining.

\section{Experiments}
\label{sec:expr}

\subsection{Datasets}
\label{sec:dataset}
We evaluate our methods on miniImageNet~\cite{russakovsky2015imagenet}, CIFAR100~\cite{krizhevsky2009learning}, and Omniglot~\cite{LakeOmniglot2015}. 
For the evaluation on miniImageNet and CIFAR100, we follow the same FSCIL procedure as in~\cite{FSCIL_CVPR2020}, dividing the dataset into a base session with 60 classes and eight novel sessions with a 5-way 5-shot problem each.
For Omniglot, we propose a new split that follows the common practice in FSCIL~\cite{FSCIL_CVPR2020}.
It contains a base session with 1200 classes and nine novel sessions with a 47-way 5-shot problem each, yielding 423 novel classes overall.
See Appendix~\ref{appendix:dataset} for more details. 

\begin{table*}[h!]
\centering
\caption{Classification accuracy (\%) on miniImageNet in the 5-way 5-shot FSCIL setting. [$*$]: Upper bound based on the visual illustration in the corresponding work. }\label{tab:miniimagenet}
\begin{tabular}{lccccccccc}
\toprule
Session ($s$)                       & 1     & 2     & 3     & 4     & 5     & 6     & 7     & 8    & 9                      \\
No. of classes $|\tilde{\mathcal{C}}^{(s)}|$                     & 60     & 65     & 70    & 75     & 80     & 85     & 90     & 95    & 100                          \\
\cmidrule(r){1-10}
AL-MML~\cite{FSCIL_CVPR2020}    & 61.31 & 50.09 & 45.17 & 41.16 & 37.48 & 35.52 & 32.19 & 29.46 & 24.42             \\
IDLVQ-C~\cite{VecQ_FSCIL_ICLR2021} & 64.77 & 59.87 & 55.93 & 52.62 & 49.88 & 47.55 & 44.83 & 43.14 & 41.84\\
Semantic KD$^{*}$~\cite{Semantic_FSCIL_CVPR2021} & $<$62 & $<$59 & $<$54 & $<$50 & $<$49 & $<$45 & $<$42 & $<$40 & $<$39\\
VAE$^{*}$~\cite{Synthesized_FSCIL_ICCV2021} & $<$62 & $<$60 & $<$54 & $<$52 & $<$50 & $<$49 & $<$46 & $<$44 & $<$43\\  
F2M~\cite{FlatMin_FSCIL_NIPS2021} & 67.28 & 63.80 & 60.38 & 57.06 & 54.08 & 51.39 & 48.82 & 46.58 & 44.65\\ 
CEC~\cite{Evolv_FSCIL_CVPR2021}                                         & 72.00 & 66.83 & 62.97 & 59.43 & 56.70 & 53.73 & 51.19 & 49.24 & 47.63               \\
\cmidrule(r){1-10}
\textbf{\name Mode\,1 $d$=512 (ours)}                                     & 76.37 & 70.94 & 66.36 & 62.64 & 59.31 & 56.02 & 53.14 & 51.04 & 48.87                \\
\textbf{\name Mode\,2 $d$=512 (ours)}                        & \textbf{76.45} & \textbf{71.23} & \textbf{66.71} & 63.01 & 60.09 & 56.73 & 53.94 & 52.01 & 50.08 \\
\textbf{\name Mode\,3 $d$=512 (ours)}                        & 76.40 & 71.14 & 66.46 & \textbf{63.29} & \textbf{60.42 }& \textbf{57.46} & \textbf{54.78} & \textbf{53.11} & \textbf{51.41} \\
\bottomrule
\end{tabular}
\end{table*}

\begin{table*}[h!]
\centering
\caption{Classification accuracy (\%) on CIFAR100 in the 5-way 5-shot FSCIL setting. [$*$]: Upper bound based on the visual illustration in the corresponding work. }\label{tab:cifar100}
\begin{tabular}{lccccccccc}
\toprule
Session ($s$)                       & 1     & 2     & 3     & 4     & 5     & 6     & 7     & 8    & 9                      \\
No. of classes $|\tilde{\mathcal{C}}^{(s)}|$                     & 60     & 65     & 70    & 75     & 80     & 85     & 90     & 95    & 100                          \\
\cmidrule(r){1-10}
AL-MML~\cite{FSCIL_CVPR2020}    & 64.10 & 55.88 & 47.07 & 45.16 & 40.11 & 36.38 & 33.96 & 31.55 & 29.37           \\
Semantic KD$^{*}$~\cite{Semantic_FSCIL_CVPR2021} & $<$64 & $<$57 & $<$51 & $<$46 & $<$43 & $<$41 & $<$39 & $<$37 & $<$35   \\
VAE$^{*}$~\cite{Synthesized_FSCIL_ICCV2021} & $<$62 & $<$58 & $<$57 &$<$52 &$<$51 & $<$49 & $<$46 & $<$45 & $<$42\\ 
F2M~\cite{FlatMin_FSCIL_NIPS2021} & 64.71 & 62.05 & 59.01 & 55.58 & 52.55 & 49.96 & 48.08 & 46.67 & 44.67 \\ 
CEC~\cite{Evolv_FSCIL_CVPR2021}   & 73.07 & 68.88 & 65.26 & 61.19 & 58.09 & 55.57 & 53.22 & 51.34 & 49.14     \\
\cmidrule(r){1-10}
\textbf{\name Mode\,1 $d$=512 (ours) }           & 77.47 & 72.20 & 67.53 & 63.23 & 59.58 & 56.67 & 53.94 & 51.55 & 49.36     \\ 
\textbf{\name Mode\,2 $d$=512 (ours)}                                     & \textbf{77.50} & \textbf{72.45} & \textbf{67.94} & \textbf{63.80} & \textbf{60.24} & \textbf{57.34} & \textbf{54.61} & 52.41 & 50.23   \\
\textbf{\name Mode\,3 $d$=512 (ours)}    & 77.47 & 72.40 & 67.47 & 63.25 & 59.84 & 56.95 & 54.42 & \textbf{52.47} & \textbf{50.47} \\
\bottomrule
\end{tabular}
\end{table*}

\begin{table*}[h!]
\centering
\caption{Classification accuracy (\%) on Omniglot in the 47-way 5-shot FSCIL setting.
[$*$]: Reproduced baselines.}\label{tab:omniglot_fscil}
\begin{tabular}{lcccccccccc}
\toprule
Session ($s$)                & 1        & 2        & 3        & 4        & 5        & 6        & 7        & 8        & 9        & 10              \\
No. of classes $|\tilde{\mathcal{C}}^{(s)}|$ & 1200 & 1247 & 1294 & 1341 & 1388 & 1435 & 1482 & 1529 & 1576 & 1623     \\
\cmidrule(r){1-11}
ProtoNet$^{*}$\cite{snell17} &  70.61 & 70.20 & 70.01 & 69.68 & 69.48 & 68.99 & 68.74 & 68.07 & 67.60 & 67.41       \\
CEC$^{*}$\cite{Evolv_FSCIL_CVPR2021} & 78.91 & 79.07 & 78.74 & 78.60 & 77.94 & 77.55 & 77.18 & 76.77 & 76.39 & 76.11   \\
\cmidrule(r){1-11}
\textbf{\name Mode\,1 $d$=512 (ours)}        & 84.16 & 83.82 & 83.69 & 83.32 & 83.22 & 82.78 & 82.70 & 82.32 & 81.77 & 81.56    \\                  
\textbf{\name Mode\,2 $d$=512 (ours)}        & 86.87 & 86.77 & 86.57 & 86.44 & 86.40 & 86.20 & \textbf{86.25} & 85.96 & 85.63 & 85.49 \\
\textbf{\name Mode\,3 $d$=512 (ours)}        & \textbf{87.21} & \textbf{87.03} & \textbf{86.89} & \textbf{86.60} & \textbf{86.43} & \textbf{86.32} & 86.13 & \textbf{85.98} & \textbf{85.59} & \textbf{85.70} \\
\bottomrule
\end{tabular}
\end{table*}

\subsection{Experimental Setup}
\paragraph{miniImageNet and CIFAR100.}
For the natural image datasets, we use a Resnet-12 architecture as feature extractor~\cite{zhang2020deepemd,DynFewSHot_CVPR2018}. 
It consists of four residual blocks with block dimensions [64, 160, 320, 640], each containing three convolutional layers with batchnorm and ReLU activation. 
After the final global average pooling, we get output activations of dimension $d_{f}=640$. 
The fully connected layer has dimension $d=512$. 
%
Motivated by~\cite{RethinkFewShot_ECCV2020} to derive good visual representations in the embedding, we pretrain the feature extractor in the standard supervised classification on the base session by replacing the EM with an additional auxiliary fully connected layer of dimension $d\times 60$. 
The additional fully connected layer is discarded and replaced by the EM after the pretraining. 
This pretraining step improved the overall accuracy on the base session by up to 15\%.
Next, the meta-learning is attained by drawing a new 60-way 5-shot problem in every iteration and updating the model based on ten queries per way. 
The model is trained for 70,000 iterations using a stochastic gradient descent (SGD) with momentum 0.9 and weight decay $5\times 10^{-4}$. 
The learning rate is initially set to $0.01$ and reduced by 10$\times$ at iterations 30,000 and 60,000.

In Mode\,2, the fully connected layer is retrained for $T=10$ iterations at an update rate $\beta=0.01$. 
In Mode\,3, the prototype nudging is done for $U=100$ iterations at an update rate $\gamma=0.01$, and the fully connected layer is retrained for $T=50$ iterations at an update rate $\beta=0.01$.

\paragraph{Omniglot.}
For the Omniglot dataset, we use a feature extractor which involves 4 convolutional layers with 128 channels and $2\times2$ maxpooling at the end of the second and fourth layer, followed by a fully connected layer that resizes the flattened embedding to $d_f=512$ before feeding to the retrainable fully connected layer ($d=512$) that outputs the prototypes.
During the meta-learning the model is trained for 70,000 iterations with an Adam optimizer with a learning rate of $10^{-4}$. 
In Mode\,2, the fully connected layer is retrained for $T=20$ iterations with a learning rate $\beta=10^{-4}$. 
In Mode\,3, the prototype nudging is done for $U=20$ iterations with an update rate $\gamma = 0.01$, followed by a similar setting for retraining the final fully connected layer.

\subsection{Comparative Results}
\paragraph{miniImageNet and CIFAR100.} 
We compare our performance on the two natural image datasets with different state-of-the-art methods~\cite{FSCIL_CVPR2020,VecQ_FSCIL_ICLR2021,FlatMin_FSCIL_NIPS2021,Evolv_FSCIL_CVPR2021,Synthesized_FSCIL_ICCV2021,Semantic_FSCIL_CVPR2021}, as shown in Table~\ref{tab:miniimagenet} and Table~\ref{tab:cifar100}. 
%
%
Our method sets the new state-of-the-art on both datasets. 
Notably, even the simple prototype averaging (Mode\,1) outperforms all other methods on both miniImageNet and CIFAR100.
Note that Mode\,1 does not involve any retraining or the use of auxiliary GAA memory.
The EM can also be compressed as shown in Table~\ref{tab:miniimagenet_compression}. 

In the other two modes, the prototype quasi-orthogonalization consistently improves the accuracy. 
The prototype bipolarization (Mode\,2) is more effective for a lower number of classes (sessions $s\leq 3$ on miniImagenet and sessions $s\leq 8$ on CIFAR100), whereas the prototype nudging (Mode\,3) outperforms all other methods for a large number of classes. 
Hence, the experimental results suggest that the classification of relatively simple problems (low number of ways) requires computationally cheap updates (e.g., Mode\,1 or 2) for the best performance. 
On the other hand, harder problems (large number of ways) benefit from more sophisticated update mechanisms (i.e., Mode\,3).

\paragraph{Omniglot.}
Table~\ref{tab:omniglot_fscil} compares the accuracy on the Omniglot dataset. 
As no prior works evaluated their methods on Omniglot in the FSCIL setting, we adapted ProtoNet~\cite{snell17} and CEC~\cite{Evolv_FSCIL_CVPR2021} as additional baselines. 
See Appendix~\ref{appendix:omniglot_baselines} for more details.
\name starts with 84.16\% accuracy in the base session and ends with 81.56\% in the last session using Mode\,1. 
This small accuracy drop is further reduced by the other modes. 
In Mode~3, \name achieves an accuracy of 87.21\% in the base session and of 85.70\% in session 10, outperforming both the ProtoNet and CEC baselines by a large margin of $\geq$16.99\% and $\geq$8.30\%, respectively. 

For an additional comparison on Omniglot, we consider an alternative continual incremental learning setting developed by~\cite{ecai2020}, that arranges 600 instead of the previous 423 classes in the novel sessions, but does not consider the evaluation on the base classes. 
As shown in Appendix~\ref{appendix:omniglot_alternative}, we find that all modes of \name performs consistently better than the best baseline, ANML~\cite{ecai2020}.

\subsection{Ablation study}
We conduct extensive ablation experiments on the dimension $d$, the attention function, and the feature extractor. 
Here, we list the main findings of the ablation. Detailed results and discussions are available in Appendix~\ref{appendix:ablation}. 
\newline
\textbf{Dimension.} 
We analyze the effect of reduced dimensions in Tables~\ref{tab:miniimagenet_d}--\ref{tab:omniglot_fscil_d}. 
We find that \name in Mode~3 allows to reduce the dimension below the number of classes ($d<|\tilde{\mathcal{C}}^{(S)}|$), e.g., $d=64$ for miniImageNet and CIFAR100 or $d=128$ for Omniglot, yielding marginal accuracy degradation while still outperforming all baseline methods. 
%
\newline
\textbf{Attention function.}
We compare the soft absolute~\cite{KarunaratneHDAugmented2021} attention with the exponential attention, which is commonly used in softmax.
Tables~\ref{tab:miniimagenet_attention}--\ref{tab:omniglot_attention} show the superiority of the soft absolute attention function, which particularly improves the accuracy when novel classes are encountered. 
\newline
\textbf{Feature extractor.}
Most baseline methods on miniImageNet and CIFAR100 use a ResNet-18 as feature extractor~\cite{FSCIL_CVPR2020,VecQ_FSCIL_ICLR2021,Semantic_FSCIL_CVPR2021,Synthesized_FSCIL_ICCV2021,FlatMin_FSCIL_NIPS2021}, requiring slightly fewer parameters compared to our ResNet-12 (12.4\,M vs. 11.\,2M). 
Tables~\ref{tab:miniimagenet_FE}--\ref{tab:cifar100_FE} show that a reduced ResNet-12 (8.0\,M parameters) maintains a high accuracy ($<$1\% drop) and still outperforms all baseline methods.

\section{Conclusion and Outlook}
We propose \name for few-shot class-incremental learning in which the model is either built rapidly in one pass without gradient-based updates (Mode\,1), or retrained with a small constant number of iterations (Modes\,2 and 3).
The \name memory grows at most linearly with the number of encountered classes.
The simple averaged prototypes in Mode\,1 outperform all other methods for CIFAR100, miniImageNet, and Omniglot. 
In Modes\,2 and 3, the optimization of the prototypes and the fixed-sized fully connected layer, through a maximum of 50 iterations, leads to higher accuracy (up to 4\%) when the maximum number of classes is encountered.

Moreover, simply training with class prototypes having large inter-class separation provides robustness against adversarial perturbations, without requiring any adversarial training~\cite{RepulsivePrototypes_2021}. 
In a similar vein, \name naturally pushes the meta-learned prototypes towards quasi-orthogonality. 
Furthermore, the precision of such robust prototypes can be reduced, as confirmed by the bipolarization in Mode\,2, which makes them ideal for implementation on emerging hardware technologies exploiting non-volatile memory for in-memory computation~\cite{Karunaratne2020,KarunaratneHDAugmented2021}.

\section*{Acknowledgements}
We acknowledge support from the IBM Research AI Hardware Center, and the Center for Computational Innovation at Rensselaer Polytechnic Institute for computational resources on the AiMOS Supercomputer.

{\small
\bibliographystyle{ieee_fullname}
\bibliography{bibliography}

\begin{thebibliography}{10}\itemsep=-1pt

\bibitem{Bang2021rainbow}
Jihwan Bang, Heesu Kim, YoungJoon Yoo, Jung-Woo Ha, and Jonghyun Choi.
\newblock Rainbow memory: Continual learning with a memory of diverse samples.
\newblock In {\em Proceedings of the IEEE/CVF Conference on Computer Vision and
  Pattern Recognition (CVPR)}, June 2021.

\bibitem{ecai2020}
Shawn Beaulieu, Lapo Frati, Thomas Miconi, Joel Lehman, Kenneth~O Stanley, Jeff
  Clune, and Nick Cheney.
\newblock Learning to continually learn.
\newblock In {\em European Conference on Artificial Intelligence (ECAI)}, 2020.

\bibitem{borsos2020coresets}
Zal\'{a}n Borsos, Mojmir Mutny, and Andreas Krause.
\newblock Coresets via bilevel optimization for continual learning and
  streaming.
\newblock In {\em Advances in Neural Information Processing Systems (NeurIPS)},
  2020.

\bibitem{E2E_CIL_ECCV2018}
Francisco~M. Castro, Manuel~J. Marin-Jimenez, Nicolas Guil, Cordelia Schmid,
  and Karteek Alahari.
\newblock End-to-end incremental learning.
\newblock In {\em Proceedings of the European Conference on Computer Vision
  (ECCV)}, September 2018.

\bibitem{VecQ_FSCIL_ICLR2021}
Kuilin Chen and Chi-Guhn Lee.
\newblock Incremental few-shot learning via vector quantization in deep
  embedded space.
\newblock In {\em International Conference on Learning Representations (ICLR)},
  2021.

\bibitem{Semantic_FSCIL_CVPR2021}
Ali Cheraghian, Shafin Rahman, Pengfei Fang, Soumava~Kumar Roy, Lars Petersson,
  and Mehrtash Harandi.
\newblock Semantic-aware knowledge distillation for few-shot class-incremental
  learning.
\newblock In {\em Proceedings of the IEEE/CVF Conference on Computer Vision and
  Pattern Recognition (CVPR)}, June 2021.

\bibitem{Synthesized_FSCIL_ICCV2021}
Ali Cheraghian, Shafin Rahman, Sameera Ramasinghe, Pengfei Fang, Christian
  Simon, Lars Petersson, and Mehrtash Harandi.
\newblock Synthesized feature based few-shot class-incremental learning on a
  mixture of subspaces.
\newblock In {\em Proceedings of the IEEE/CVF International Conference on
  Computer Vision (ICCV)}, 2021.

\bibitem{cong2020GANforgetting}
Yulai Cong, Miaoyun Zhao, Jianqiao Li, Sijia Wang, and Lawrence Carin.
\newblock Gan memory with no forgetting.
\newblock In {\em Advances in Neural Information Processing Systems (NeurIPS)},
  2020.

\bibitem{frady2020resonator}
E~Paxon Frady, Spencer~J Kent, Bruno~A Olshausen, and Friedrich~T Sommer.
\newblock Resonator networks, 1: An efficient solution for factoring
  high-dimensional, distributed representations of data structures.
\newblock {\em Neural computation}, 32(12):2311--2331, 2020.

\bibitem{VSA03}
Ross~W. Gayler.
\newblock Vector symbolic architectures answer {Jackendoff's} challenges for
  cognitive neuroscience.
\newblock In {\em {Proceedings of the Joint International Conference on
  Cognitive Science. ICCS/ASCS}}, pages 133--138, 2003.

\bibitem{DynFewSHot_CVPR2018}
Spyros Gidaris and Nikos Komodakis.
\newblock Dynamic few-shot visual learning without forgetting.
\newblock In {\em Proceedings of the IEEE Conference on Computer Vision and
  Pattern Recognition (CVPR)}, June 2018.

\bibitem{catastrophicForgetting_ICLR14}
Ian~J. Goodfellow, Mehdi Mirza, Da Xiao, Aaron Courville, and Yoshua Bengio.
\newblock An empirical investigation of catastrophic forgeting in
  gradient-based neural networks.
\newblock In {\em Proceedings of International Conference on Learning
  Representations (ICLR)}, 2014.

\bibitem{graves14}
Alex Graves, Greg Wayne, and Ivo Danihelka.
\newblock {Neural Turing Machines}.
\newblock {\em arXiv:1410.5401}, 2014.

\bibitem{graves16}
Alex Graves, Greg Wayne, Malcolm Reynolds, Tim Harley, Ivo Danihelka, Agnieszka
  Grabska-Barwi{\'n}ska, Sergio~G{\'o}mez Colmenarejo, Edward Grefenstette,
  Tiago Ramalho, John Agapiou, Adri{\`a}~Puigdom{\`e}nech Badia, Karl~Moritz
  Hermann, Yori Zwols, Georg Ostrovski, Adam Cain, Helen King, Christopher
  Summerfield, Phil Blunsom, Koray Kavukcuoglu, and Demis Hassabis.
\newblock {Hybrid Computing Using a Neural Network with Dynamic External
  Memory}.
\newblock {\em Nature}, 538(7626):471--476, 2016.

\bibitem{ResNet_CVPR2016}
Kaiming He, Xiangyu Zhang, Shaoqing Ren, and Jian Sun.
\newblock Deep residual learning for image recognition.
\newblock In {\em Proceedings of the IEEE Conference on Computer Vision and
  Pattern Recognition (CVPR)}, June 2016.

\bibitem{hersche2022nvsa}
Michael Hersche, Mustafa Zeqiri, Luca Benini, Abu Sebastian, and Abbas Rahimi.
\newblock A neuro-vector-symbolic architecture for solving {Raven's}
  progressive matrices.
\newblock {\em arXiv:2203.04571}, 2022.

\bibitem{FixClassifier_ICLR18}
Elad Hoffer, Itay Hubara, and Daniel Soudry.
\newblock Fix your classifier: the marginal value of training the last weight
  layer.
\newblock In {\em International Conference on Learning Representations (ICLR)},
  2018.

\bibitem{Rebalancing_CIL_CVPR2019}
Saihui Hou, Xinyu Pan, Chen~Change Loy, Zilei Wang, and Dahua Lin.
\newblock Learning a unified classifier incrementally via rebalancing.
\newblock In {\em Proceedings of the IEEE/CVF Conference on Computer Vision and
  Pattern Recognition (CVPR)}, June 2019.

\bibitem{KanervaHyperdimensional2009}
P. Kanerva.
\newblock Hyperdimensional computing: An introduction to computing in
  distributed representation with high-dimensional random vectors.
\newblock {\em Cognitive Computation}, 1(2):139--159, 2009.

\bibitem{Karunaratne2020}
Geethan Karunaratne, Manuel Le~Gallo, Giovanni Cherubini, Luca Benini, Abbas
  Rahimi, and Abu Sebastian.
\newblock {In-memory Hyperdimensional Computing}.
\newblock {\em {Nature Electronics}}, 3(6):327--337, 2020.

\bibitem{KarunaratneHDAugmented2021}
Geethan Karunaratne, Manuel Schmuck, Manuel Le~Gallo, Giovanni Cherubini, Luca
  Benini, Abu Sebastian, and Abbas Rahimi.
\newblock {Robust High-dimensional Memory-augmented Neural Networks}.
\newblock {\em {Nature Communications}}, 12(1):1--12, 2021.

\bibitem{Regular_CatastForget_2017}
James Kirkpatrick, Razvan Pascanu, Neil Rabinowitz, Joel Veness, Guillaume
  Desjardins, Andrei~A. Rusu, Kieran Milan, John Quan, Tiago Ramalho, Agnieszka
  Grabska-Barwinska, Demis Hassabis, Claudia Clopath, Dharshan Kumaran, and
  Raia Hadsell.
\newblock Overcoming catastrophic forgetting in neural networks.
\newblock {\em Proceedings of the National Academy of Sciences},
  114(13):3521--3526, 2017.

\bibitem{GenKV_TNNLS22}
Denis Kleyko, Geethan Karunaratne, Jan Rabaey, Abu Sebastian, and Abbas Rahimi.
\newblock {Generalized Key-Value Memory to Flexibly Adjust Redundancy in
  Memory-Augmented Networks}.
\newblock {\em {IEEE Transactions on Neural Networks and Learning Systems}},
  2022.

\bibitem{BiT_ECCV2020}
Alexander Kolesnikov, Lucas Beyer, Xiaohua Zhai, Joan Puigcerver, Jessica Yung,
  Sylvain Gelly, and Neil Houlsby.
\newblock Big transfer (bit): General visual representation learning.
\newblock In {\em Proceedings of the European Conference on Computer Vision
  (ECCV)}, 2020.

\bibitem{krizhevsky2009learning}
Alex Krizhevsky.
\newblock Learning multiple layers of features from tiny images.
\newblock {\em University of Toronto}, 2009.

\bibitem{AlexNet_NIPS2012}
Alex Krizhevsky, Ilya Sutskever, and Geoffrey~E Hinton.
\newblock Imagenet classification with deep convolutional neural networks.
\newblock In {\em Advances in Neural Information Processing Systems (NeurIPS)},
  2012.

\bibitem{LakeOmniglot2015}
B.~M. Lake, R. Salakhutdinov, and J.~B. Tenenbaum.
\newblock Human-level concept learning through probabilistic program induction.
\newblock {\em {Science}}, 350:1332--1338, 2015.

\bibitem{LedouxConcentration2001}
M. Ledoux.
\newblock {\em {The Concentration of Measure Phenomenon}}.
\newblock American Mathematical Society, 2001.

\bibitem{LastLayer_CIL_2021}
Timothée Lesort, Thomas George, and Irina Rish.
\newblock Continual learning in deep networks: an analysis of the last layer.
\newblock {\em arXiv:2106.01834}, 2021.

\bibitem{Regular_CIL_2021}
Timothée Lesort, Andrei Stoian, and David Filliat.
\newblock Regularization shortcomings for continual learning.
\newblock {\em arXiv:1912.03049}, 2021.

\bibitem{Liu_2020_CVPR}
Yaoyao Liu, Yuting Su, An-An Liu, Bernt Schiele, and Qianru Sun.
\newblock Mnemonics training: Multi-class incremental learning without
  forgetting.
\newblock In {\em Proceedings of the IEEE/CVF Conference on Computer Vision and
  Pattern Recognition (CVPR)}, June 2020.

\bibitem{GradientEM_NIPS17}
David Lopez-Paz and Marc'Aurelio Ranzato.
\newblock Gradient episodic memory for continual learning.
\newblock In {\em Advances in Neural Information Processing Systems (NeurIPS)},
  2017.

\bibitem{catastrophicForgetting_1989}
Michael McCloskey and Neal~J. Cohen.
\newblock Catastrophic interference in connectionist networks: The sequential
  learning problem.
\newblock volume~24 of {\em Psychology of Learning and Motivation}, pages
  109--165. Academic Press, 1989.

\bibitem{LatenReplay_IROS2020}
Lorenzo Pellegrini, Gabriele Graffieti, Vincenzo Lomonaco, and Davide Maltoni.
\newblock Latent replay for real-time continual learning.
\newblock In {\em IEEE/RSJ International Conference on Intelligent Robots and
  Systems (IROS)}, October 2020.

\bibitem{Plate_TNN95}
T.~A. {Plate}.
\newblock Holographic reduced representations.
\newblock {\em IEEE Transactions on Neural Networks}, 6(3):623--641, 1995.

\bibitem{prabhu2020gdumb}
Ameya Prabhu, Philip~HS Torr, and Puneet~K Dokania.
\newblock Gdumb: A simple approach that questions our progress in continual
  learning.
\newblock In {\em Proceedings of the European Conference on Computer Vision
  (ECCV)}, 2020.

\bibitem{IdentityMatx_2020}
Zhongchao Qian, Tyler~L. Hayes, Kushal Kafle, and Christopher Kanan.
\newblock Do we need fully connected output layers in convolutional networks?
\newblock {\em arXiv:2004.13587}, 2020.

\bibitem{PULP_LR_2021}
Leonardo Ravaglia, Manuele Rusci, Davide Nadalini, Alessandro Capotondi,
  Francesco Conti, and Luca Benini.
\newblock A tinyml platform for on-device continual learning with quantized
  latent replays.
\newblock {\em IEEE Journal on Emerging and Selected Topics in Circuits and
  Systems}, 2021.

\bibitem{iCaRL_CVPR2017}
Sylvestre-Alvise Rebuffi, Alexander Kolesnikov, Georg Sperl, and Christoph~H.
  Lampert.
\newblock icarl: Incremental classifier and representation learning.
\newblock In {\em Proceedings of the IEEE Conference on Computer Vision and
  Pattern Recognition (CVPR)}, July 2017.

\bibitem{AttAttrac_FSCIL_NIPS19}
Mengye Ren, Renjie Liao, Ethan Fetaya, and Richard~S. Zemel.
\newblock Incremental few-shot learning with attention attractor networks.
\newblock In {\em Advances in Neural Information Processing Systems (NeurIPS)},
  2019.

\bibitem{russakovsky2015imagenet}
Olga Russakovsky, Jia Deng, Hao Su, Jonathan Krause, Sanjeev Satheesh, Sean Ma,
  Zhiheng Huang, Andrej Karpathy, Aditya Khosla, Michael Bernstein, et~al.
\newblock Imagenet large scale visual recognition challenge.
\newblock {\em International journal of computer vision}, 115(3):211--252,
  2015.

\bibitem{MANN_pmlr2016}
Adam Santoro, Sergey Bartunov, Matthew Botvinick, Daan Wierstra, and Timothy
  Lillicrap.
\newblock Meta-learning with memory-augmented neural networks.
\newblock In {\em Proceedings of The 33rd International Conference on Machine
  Learning}. PMLR, 2016.

\bibitem{RepulsivePrototypes_2021}
Alex Serban, Erik Poll, and Joost Visser.
\newblock Deep repulsive prototypes for adversarial robustness.
\newblock {\em arXiv:2105.12427}, 2021.

\bibitem{FlatMin_FSCIL_NIPS2021}
Guangyuan Shi, Jiaxin Chen, Wenlong Zhang, Li-Ming Zhan, and Xiao-Ming Wu.
\newblock Overcoming catastrophic forgetting in incremental few-shot learning
  by finding flat minima.
\newblock In {\em Advances in Neural Information Processing Systems (NeurIPS)},
  2021.

\bibitem{snell17}
Jake Snell, Kevin Swersky, and Richard Zemel.
\newblock {Prototypical Networks for Few-Shot Learning}.
\newblock In {\em {Advances in Neural Information Processing Systems
  (NeurIPS)}}, 2017.

\bibitem{sukhbaatar15}
Sainbayar Sukhbaatar, arthur szlam, Jason Weston, and Rob Fergus.
\newblock {End-To-End Memory Networks}.
\newblock In {\em Advances in Neural Information Processing Systems (NeurIPS)},
  2015.

\bibitem{DeeperCNN_CVPR2015}
Christian Szegedy, Wei Liu, Yangqing Jia, Pierre Sermanet, Scott Reed, Dragomir
  Anguelov, Dumitru Erhan, Vincent Vanhoucke, and Andrew Rabinovich.
\newblock Going deeper with convolutions.
\newblock In {\em Proceedings of the IEEE Conference on Computer Vision and
  Pattern Recognition (CVPR)}, June 2015.

\bibitem{FSCIL_CVPR2020}
Xiaoyu Tao, Xiaopeng Hong, Xinyuan Chang, Songlin Dong, Xing Wei, and Yihong
  Gong.
\newblock Few-shot class-incremental learning.
\newblock In {\em Proceedings of the IEEE/CVF Conference on Computer Vision and
  Pattern Recognition (CVPR)}, June 2020.

\bibitem{RethinkFewShot_ECCV2020}
Yonglong Tian, Yue Wang, Dilip Krishnan, Joshua~B. Tenenbaum, and Phillip
  Isola.
\newblock Rethinking few-shot image classification: A good embedding is all you
  need?
\newblock In {\em Proceedings of the European Conference on Computer Vision
  (ECCV)}, 2020.

\bibitem{matching_net}
Oriol Vinyals, Charles Blundell, Timothy Lillicrap, Koray Kavukcuoglu, and Daan
  Wierstra.
\newblock {Matching Networks for One Shot Learning}.
\newblock In {\em {Advances in Neural Information Processing Systems
  (NeurIPS)}}, 2016.

\bibitem{weston15}
Jason Weston, Sumit Chopra, and Antoine Bordes.
\newblock {Memory Networks}.
\newblock In {\em International Conference on Learning Representations (ICLR)},
  2015.

\bibitem{LargeScl_CIL_CVPR2019}
Yue Wu, Yinpeng Chen, Lijuan Wang, Yuancheng Ye, Zicheng Liu, Yandong Guo, and
  Yun Fu.
\newblock Large scale incremental learning.
\newblock In {\em Proceedings of the IEEE/CVF Conference on Computer Vision and
  Pattern Recognition (CVPR)}, June 2019.

\bibitem{Wu_2018_ECCV}
Zhirong Wu, Alexei~A. Efros, and Stella~X. Yu.
\newblock {Improving Generalization via Scalable Neighborhood Component
  Analysis}.
\newblock In {\em Proceedings of the European Conference on Computer Vision
  (ECCV)}, 2018.

\bibitem{Wu_2018_CVPR}
Zhirong Wu, Yuanjun Xiong, Stella~X. Yu, and Dahua Lin.
\newblock {Unsupervised Feature Learning via Non-Parametric Instance
  Discrimination}.
\newblock In {\em IEEE Conference on Computer Vision and Pattern Recognition
  (CVPR)}, 2018.

\bibitem{zhang2020deepemd}
Chi Zhang, Yujun Cai, Guosheng Lin, and Chunhua Shen.
\newblock Deepemd: Few-shot image classification with differentiable earth
  mover's distance and structured classifiers.
\newblock In {\em Proceedings of the IEEE Conference on Computer Vision and
  Pattern Recognition (CVPR)}, 2020.

\bibitem{Evolv_FSCIL_CVPR2021}
Chi Zhang, Nan Song, Guosheng Lin, Yun Zheng, Pan Pan, and Yinghui Xu.
\newblock Few-shot incremental learning with continually evolved classifiers.
\newblock In {\em Proceedings of the IEEE/CVF Conference on Computer Vision and
  Pattern Recognition (CVPR)}, June 2021.

\bibitem{WeightAligning_CVPR2020}
Bowen Zhao, Xi Xiao, Guojun Gan, Bin Zhang, and Shu-Tao Xia.
\newblock Maintaining discrimination and fairness in class incremental
  learning.
\newblock In {\em Proceedings of the IEEE/CVF Conference on Computer Vision and
  Pattern Recognition (CVPR)}, June 2020.

\end{thebibliography}
}

\clearpage
\appendix 
\setcounter{figure}{0}
\renewcommand{\thefigure}{A\arabic{figure}}
\setcounter{table}{0}
\renewcommand{\thetable}{A\arabic{table}}

\section{Appendix}

\subsection{Datasets}\label{appendix:dataset}
\paragraph{miniImageNet.} 
miniImageNet contains RGB images of size 84$\times$84 from 100 different classes, where each class has 500 training images and 100 test images.  
It is a subset of the much larger ImageNet dataset~\cite{russakovsky2015imagenet} and was initially proposed for few-shot learning problems~\cite{matching_net}.
For the FSCIL evaluation, we follow the same procedure as in~\cite{FSCIL_CVPR2020}, dividing the dataset into a base session with 60 classes and eight novel sessions with a 5-way 5-shot problem each. 

\paragraph{CIFAR100.} 
The setup of CIFAR100~\cite{krizhevsky2009learning} is similar to miniImageNet, whereby CIFAR contains 100 different classes with 500 training images and 100 testing images per class. 
The resolution of the images is lower (32$\times$32). 
Also here, we follow the same FSCIL procedure as in~\cite{FSCIL_CVPR2020} with 60 base classes and eight novel sessions with 5-way 5-shot problems.

\paragraph{Omniglot.}
The Omniglot dataset~\cite{LakeOmniglot2015} has a total of 1623 classes with 20 example images each.
It is publicly available under the MIT license.  
The images are binary with a size of 105$\times$105. 
We resize all the images to 32$\times$32 floating point format in a preprocessing step. 
As proposed by Vinyals et al. in~\cite{matching_net} for the few-shot learning setting, we use 1200 base classes for the meta-learning and base session while the remaining 423 classes are reserved as the novel classes for the subsequent sessions. 
However, as there has been no previous work targeting Omniglot in the FSCIL setting, we consider the following points for the dataset.

First, to accommodate evaluation within the base classes, we hold out the last 6 samples from the base classes, leaving the first 14 samples for the training dataset. 
 %
%
Second, we demarcate the first 5 samples from the next-in-line 47 novel classes as the incoming support batch during sessions subsequent to the base session, so that 9 subsequent sessions can be run with 1623 classes in total in the final session. 
Third, we add the first 6 of the remaining 15 examples from the novel classes for the evaluation query batch, so that novel and base classes are equally weighted during the evaluation.

\subsection{Ablation study}\label{appendix:ablation}

\subsubsection{Reducing Dimension}
\label{sec:dim}
We analyse the classification accuracy by reducing the dimension $d \in \{ 32, 64, 128, 256, 512\}$ of the output of the fully connected layer and the EM for the three datasets, whereby the maximum number of classes ($|\tilde{\mathcal{C}}^{(S)}|$) is set to 100, 100, and 1623 in CIFAR100, miniImageNet, and Omniglot, respectively. 
Experimental results are shown in Table~\ref{tab:miniimagenet_d}, Table~\ref{tab:cifar100_d}, and Table~\ref{tab:omniglot_fscil_d}. 
All training hyperparamters applied in the meta-learning and retraining are kept the same, irrespective of the dimensionality. 
Overall, in Mode~1 and Mode~2, high dimensionality, i.e., $d \ge 256$, leads to better accuracy.
This stems from the fact that these two modes mainly rely on the property of hyperdimensional vectors---where higher dimensionality is preferred---to achieve quasi-orthogonality between class vectors.
However, the optimization technique employed in Mode~3 is able to find a better distribution of the prototypes with a lower dimensionality of $d=128 > |\tilde{\mathcal{C}}^{(S)}|$ for CIFAR and miniImageNet datasets. 
Specifically, we see the advantage of a lower number of dimensions when increasing the number of novel classes, provided the number of dimensions is larger than the total number of classes. 
For example, in the last session, \name with $d=128$ achieves the highest accuracy on both miniImageNet (51.46\%) and CIFAR100 (50.74\%) in Mode~3.
We could not observe this effect in Omniglot (Table~\ref{tab:omniglot_fscil_d}), where the highest dimensionality is lower than the number of classes (i.e., $d=512<|\tilde{\mathcal{C}}^{(S)}|$). 
In fact, $d=512$ results in the highest accuracy for all the modes in Omniglot.

We remark that, compared to the state-of-the-art, the superior accuracy of \name is still maintained even with $d<|\tilde{\mathcal{C}}^{(S)}|$: 
\begin{itemize}
    \item On miniImageNet, \name with $d=64$ outperforms~\cite{FSCIL_CVPR2020,VecQ_FSCIL_ICLR2021,Semantic_FSCIL_CVPR2021,Synthesized_FSCIL_ICCV2021,FlatMin_FSCIL_NIPS2021, Evolv_FSCIL_CVPR2021} in Mode~1 and Mode~3, and ~\cite{FSCIL_CVPR2020,VecQ_FSCIL_ICLR2021,Semantic_FSCIL_CVPR2021,Synthesized_FSCIL_ICCV2021,FlatMin_FSCIL_NIPS2021} in Mode~2.
    
    \item Similarly, on CIFAR100, \name with $d=64$ outperforms~\cite{FSCIL_CVPR2020,VecQ_FSCIL_ICLR2021,Semantic_FSCIL_CVPR2021,Synthesized_FSCIL_ICCV2021,FlatMin_FSCIL_NIPS2021} in Mode~1 and Mode~2, and~\cite{FSCIL_CVPR2020,VecQ_FSCIL_ICLR2021,Semantic_FSCIL_CVPR2021,Synthesized_FSCIL_ICCV2021,FlatMin_FSCIL_NIPS2021, Evolv_FSCIL_CVPR2021} in Mode~3.
    
    \item Likewise, on Omniglot, \name with $d=128$ in any mode outperforms the prototypical network~\cite{snell17} and CEC~\cite{Evolv_FSCIL_CVPR2021}. 
\end{itemize}

\subsubsection{Other Attention Functions}
\label{sec:loss}
In this section, we provide additional details on the soft absolute (softabs) attention function applied in our meta-learning, and compare it to the exponential attention commonly used in the softmax. 

\paragraph{Softabs attention.} 
Given the cosine similarity score $l_j$ for every class $j$ (see~\eqref{eq:score} in the main paper), the softabs attention function is defined as 
\begin{align}
    h(l_j) = \frac{\epsilon(l_j)}{\sum_{i=1}^{|\tilde{\mathcal{C}}^{(s)}|}  \epsilon(l_i)} , 
\end{align}
where $\epsilon(\cdot)$ is the sharpening function: 
\begin{align}
    \epsilon(c) = \frac{1}{1+e^{-(\beta (c-0.5))}} + \frac{1}{1+e^{-(\beta (-c-0.5))}}, \quad \forall c \in \Reals. 
\end{align}
The sharpening function includes a stiffness parameter $\beta$, which is set to 10 as in~\cite{KarunaratneHDAugmented2021}.
During meta-learning, the model is updated based on the negative log-likelihood loss applied on $h(l_{y})$. 
The sharpening function is maximized at $c=1$ or $c=-1$, and minimized at $c=0$. 
Hence, it promotes orthogonal prototypes. 
This notion of orthogonality is also reflected in the activation function applied in the prototype nudging in Mode~3 (see~\eqref{eq:doubleexp} in the main paper):
\begin{align}\label{eq:doubleexp_nuding_act}
    \sigma(c) := e^{\alpha \cdot c}+ e^{- \alpha \cdot c} -2, \forall c\in \Reals,
\end{align} 
where $\alpha=4$. This activation function penalizes the prototype pairs with large absolute cross-correlations. 

\paragraph{Softmax attention.}
We compare the aforementioned softabs attention with the conventional exponential softmax, defined as
\begin{align} \label{eq:softmax_sharp}
    r(l_j) = \frac{e^{\tau \cdot l_j}}{\sum_{i=1}^{|\tilde{\mathcal{C}}^{(s)}|} e^{ \tau \cdot l_i}}, 
\end{align}
where $\tau=10$ is the inverse softmax temperature.
When applying the negative log-likelihood loss on $r(l_y)$, we get the commonly used categorical cross-entropy loss (CEL). 
The CEL aims to find anti-correlating prototypes.  
During prototype nudging in Mode~3, we therefore modify the activation function~\eqref{eq:doubleexp_nuding_act} with the objective of reaching anti-correlation: 
\begin{align}\label{eq:exp_nudging_act}
    \sigma'(c) := e^{\alpha \cdot c} -1, \forall c\in \Reals,
\end{align}
where $\alpha=4$ as in~\eqref{eq:doubleexp_nuding_act}. 

\paragraph{Comparison.}
We compare the classification accuracy when using either softmax or softabs attention on miniImageNet (Table~\ref{tab:miniimagenet_attention}), CIFAR100 (Table~\ref{tab:cifar100_attention}), and Omniglot (Table~\ref{tab:omniglot_attention}).
In the case of the softmax attention, we also applied pretraining of the embedding and optimized the inverse softmax temperature with a grid-search. 
On miniImageNet, the softmax attention starts with marginally higher accuracy (0.1\%) than the softabs attention in the base session, but it decays faster than the softabs when new sessions are added, independent of the mode. 
As a result, the softabs attention maintains higher accuracy, as high as 1.36\%, during the novel sessions ($s>1$). 
Similar results are observed on CIFAR100, where softabs outperforms softmax (up to 2.66\%) in all sessions and all Modes~1--3. 
Similar results are also observed by using Mode~2 and Mode~3 on Omniglot; however, the softmax attention reaches consistently higher accuracy than softabs in Mode~1.

\begin{figure*}[hb!]
    \centering
    \def\svgwidth{\textwidth}
    \fontsize{7}{9}
    \selectfont
    \includegraphics[width=\textwidth]{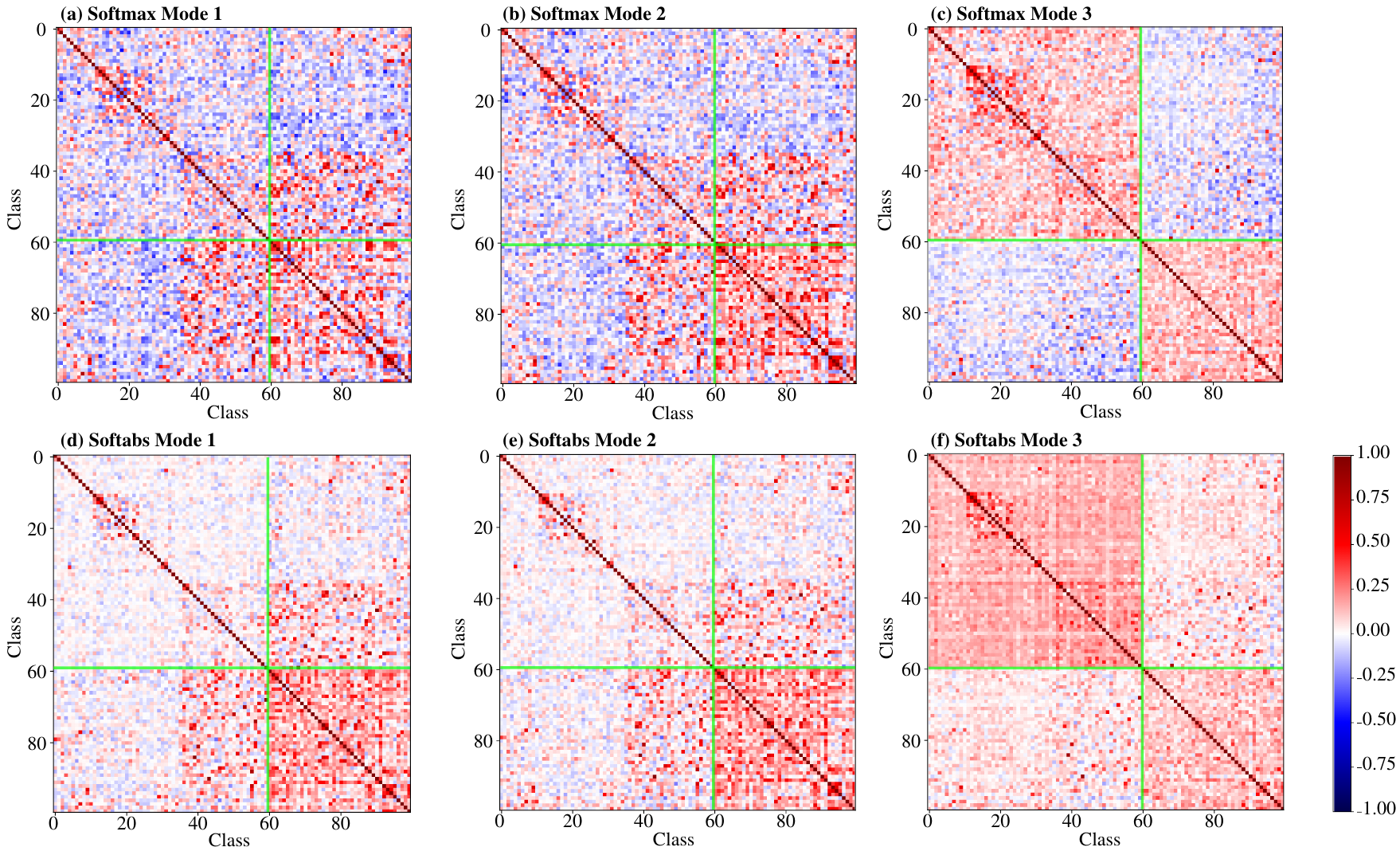}
    \caption{Cosine similarities between the prototypes on miniImageNet using different attention functions (softmax vs softabs) across the three modes. 
    The green cross splits the base session (60 classes) and the novel sessions (40 classes in total).   
    }
    \label{fig:correlation}
\end{figure*}

\begin{figure*}[hb!]
    \centering
    \def\svgwidth{\textwidth}
    \fontsize{7}{9}
    \selectfont
    \includegraphics[width=\textwidth]{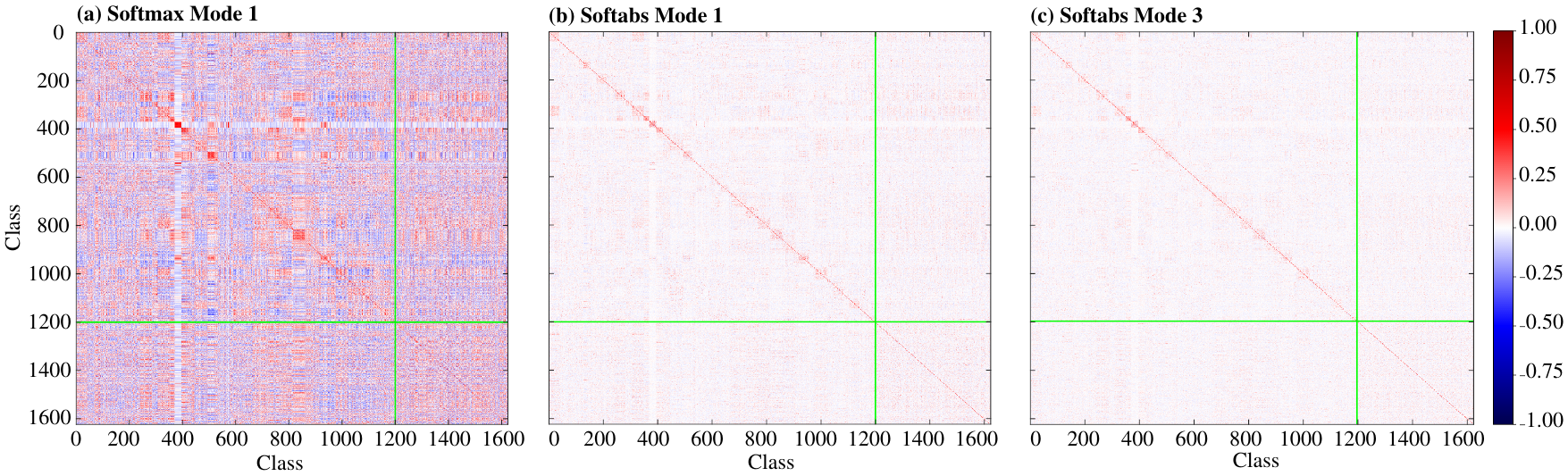}
    \caption{Cosine similarities between the prototypes on Omniglot in Mode~1 using different attention functions (softmax vs softabs). It also compares Mode~1 and Mode~3 when using softabs attention function. 
    The green cross splits the base session (1200 classes) and the novel sessions (423 classes in total).   
    }
    \label{fig:correlation_omniglot}
\end{figure*}

Fig.~\ref{fig:correlation} illustrates the relations between the prototypes, either trained with the softabs or the softmax attention. 
When comparing the softmax with the softabs in Mode~1 (Fig.~\ref{fig:correlation}a vs. Fig.~\ref{fig:correlation}d) on the base session, the softabs attention yields cross-correlations that are close to zero (i.e., they are quasi-orthogonal), whereas the softmax promotes anti-correlating prototypes with negative cross-correlations. 
When new sessions are added, both attention functions yield cross-talk in Mode~1.
This cross-talk is effectively reduced with the prototype nudging and retraining applied in Mode~3, where the exponential-based nudging in~\eqref{eq:exp_nudging_act} (Fig.~\ref{fig:correlation}c) yields anti-correlating prototypes between the novel and the base session, whereas the double-exponential nudging in~\eqref{eq:doubleexp_nuding_act} (Fig.~\ref{fig:correlation}f) yields quasi-orthogonal prototypes. 
While the cross-talk is reduced on the novel classes with both activation functions, part of the class discriminability achieved during the base session is sacrificed.
Fig.~\ref{fig:correlation_omniglot} shows similar trends on the Omniglot dataset. 
\name using the softabs in Mode~1 (Fig.~\ref{fig:correlation_omniglot}b) yields lower cross-correlations than softmax in Mode~1 (Fig.~\ref{fig:correlation_omniglot}). 
The retraining in Mode~3 further reduces the cross-talk (Fig.~\ref{fig:correlation_omniglot}c).

\subsubsection{Smaller Feature Extractor}
\label{sec:FE}
Most of the baseline methods~\cite{FSCIL_CVPR2020,VecQ_FSCIL_ICLR2021,Semantic_FSCIL_CVPR2021,Synthesized_FSCIL_ICCV2021,FlatMin_FSCIL_NIPS2021} use a ResNet-18 backbone with the feature dimensionality $d_f=512$, while we use a ResNet-12 as the feature extractor with $d_f=640$, motivated by~\cite{zhang2020deepemd,DynFewSHot_CVPR2018}. 
This higher feature dimensionality requires a larger number of trainable parameters\footnote{Final fully connected layer excluded.} for ResNet-12 compared ResNet-18 (12.4\,M vs. 11.2\,M). 
Therefore, to have a fair comparison, we have also implemented a reduced ResNet-12 feature extractor with the block dimensions [64, 128, 256, $d_f=512$] containing 8.0\,M parameters.
We name this smaller feature extractor as ResNet-12 (small).
It results in 1.56$\times$ lower number of trainable parameters than ResNet-18.

Table~\ref{tab:miniimagenet_FE} and Table~\ref{tab:cifar100_FE} compare the performance on miniImageNet and CIFAR100, respectively, applying \name with either the ResNet-12 (small) or the original ResNet-12 with $d_f=640$. 
\name using the ResNet-12 (small) maintains a high accuracy on both datasets and shows only small drops ($<$1\%) compared to the original ResNet-12 with $d_f=640$. 
Moreover, \name with ResNet-12 (small) outperforms all the baselines~\cite{FSCIL_CVPR2020,VecQ_FSCIL_ICLR2021,Semantic_FSCIL_CVPR2021,Synthesized_FSCIL_ICCV2021,FlatMin_FSCIL_NIPS2021, Evolv_FSCIL_CVPR2021} on miniImageNet in all Modes~1--3, while requiring a lower number of trainable parameters. 
Similarly, on CIFAR100, \name with ResNet-12 (small) outperforms the majority of baselines~\cite{FSCIL_CVPR2020,VecQ_FSCIL_ICLR2021,Semantic_FSCIL_CVPR2021,Synthesized_FSCIL_ICCV2021,FlatMin_FSCIL_NIPS2021} in Modes~1--3 with a lower number of trainable parameters. 
When comparing to CEC~\cite{Evolv_FSCIL_CVPR2021}, \name with ResNet-12 (small) achieves a higher accuracy in Mode~2 and Mode~3. However, we observe that CEC uses ResNet-20, which requires a lower number of parameters.

\subsection{Compression of the Explicit and GAA Memories}\label{appendix:compression}
Here, we present a case where the memory requirements of our \name can be further reduced by doing superposition of key-value bindings using holographic reduced representations~\cite{Plate_TNN95}. 
We bind each prototype with a randomly drawn key and superimpose two key-prototype pairs, which compresses the memory by 2$\times$. 
More formally, the first two prototypes, $\mathbf{p}_1$ and $\mathbf{p}_2$, are compressed by 
\begin{align}
    \mathbf{r} = \mathbf{p}_1\varoast \mathbf{c}_1 + \mathbf{p}_2\varoast \mathbf{c}_2, 
\end{align}
where $\mathbf{c}_1$ and $\mathbf{c}_2$ are $d$-dimensional key-vectors randomly drawn from a normal distribution with variance 1/$d$, and $\varoast$ is the circular convolution acting as binding operator. 
The keys are generated with a pseudorandom number generator (RG) with seed corresponding to key$_i$. 
We need to store only the seed key$_i$ instead of the actual key vector.
This key$_i$ needs a negligible 32-bit storage per model since $\mathbf{c}_i$ can be reproduced from the key by RG.
The key-value binding allows to retrieve the individual prototypes using the unbinding operation, e.g., the first prototype is retrieved by: 
\begin{align}
    \mathbf{\hat{p}}_1 &= \mathbf{r} \odot \mathbf{c}_1 \\
                        &= \mathbf{p}_1\varoast \mathbf{c}_1\odot \mathbf{c}_1  + \mathbf{p}_2\varoast \mathbf{c}_2\odot \mathbf{c}_1 \\
                        &\approx \mathbf{p}_1 + \mathbf{n}
\end{align}
where $\odot$ is the circular correlation and $\mathbf{n}$ a noise term, which decreases with increasing dimension $d$~\cite{Plate_TNN95}. 
The presented compression can be applied in all modes. 
In Mode~1, the prototype vectors in the EM are compressed, whereas in Mode~3 the globally average activation vectors in the GAA memory are compressed. 

Table~\ref{tab:miniimagenet_compression} compares the accuracy of \name with and without memory compression on miniImageNet. 
The compressed EM in Mode\,1 remains accurate (1.7\%--3.5\%\,drop across the sessions), while the compressed GAA memory in Mode\,3 yielded a larger loss (4.7\%--8.5\% drop). 
The superior accuracy of the compressed EM compared to the compressed GAA memory might stem from its quasi-orthogonal representation, which is not provided by the GAA memory. 

\subsection{Additional Baselines on Omniglot}
\label{appendix:omniglot_baselines}
For further comparison with the Omniglot dataset in the FSCIL setting, we create two new baselines based on Prototypical Networks and Continually Evolved Classifiers. 
For an additional comparison on Omniglot, we consider an alternative continual incremental learning setting developed by~\cite{ecai2020}.

\subsubsection{Prototypical Networks}
The first baseline adapts the loss function and sharpening function of \name to those used in Prototypical Networks~\cite{snell17}.
Therefore, we call this baseline as ProtoNet$^*$.
ProtoNet$^*$ adopts the same feature extractor as \name used for the Omniglot dataset. 
The averaged prototypes and query vector produced by the feature extractor are compared using the negative Euclidean distance metric, as suggested in~\cite{snell17}. 
This output attention vector goes through an exponential sharpening function, as given in~\eqref{eq:softmax_sharp}. 
During the meta-learning phase, the ProtoNet$^*$ feature extractor is trained by applying the cross-entropy loss (CEL) on the sharpened attention activations. %
During the inference phase for the base session and the subsequent sessions, the averaged prototypes are computed using forward propagation of support examples through the meta-learned feature extractor and averaging the resulting output embeddings. 
For the prediction, the query vector produced by the feature extractor is compared against the averaged prototypes using the negative Euclidean distance metric. 
To have a fair comparison, we also varied the number of output embedding dimensions in ProtoNet$^*$, although the original Prototypical Networks~\cite{snell17} used a fixed $d=64$. 

The resulting classification accuracy is presented in Table~\ref{tab:omniglot_fscil_d}.
\name in any mode significantly outperforms ProtoNet$^*$ with the same $d$.  
For instance, with $d=128$, \name starts with 17.03\% higher accuracy (80.78\% vs. 63.75\%) in the base session, and ends with 18.06\% higher accuracy in the last session using Mode~1, which is similar to the prototype averaging applied in ProtoNet$^*$.
These accuracy gaps become larger by using either Mode~2 or Mode~3.

\subsubsection{Continually Evolved Classifiers}
The second baseline is the Continually Evolved Classifiers (CEC)~\cite{Evolv_FSCIL_CVPR2021}, which achieved state-of-the-art accuracy on miniImageNet and CIFAR100 in FSCIL. 
After evaluating the performance of CEC with different feature extractors, including a ResNet-18, a ResNet-20, and the feature extractor from our C-FSCIL, we found the ResNet-20 to achieve the highest accuracy. 
Table~\ref{tab:omniglot_fscil_d} shows the accuracy when varying the embedding dimension. 
CEC achieved highest accuracy when the dimension is set to $d=64$, which is indeed the default dimension of CEC. 
Overall, our \name in Mode~3 outperforms the CEC baseline by a large margin of 8.30\% and 9.59\% in session 1 and 10, respectively. 

\subsubsection{Alternative FSCIL setting on Omniglot}\label{appendix:omniglot_alternative}
We also consider an alternative continual incremental learning setting developed by~\cite{ecai2020}, that arranges a larger number of classes in the novel sessions.
In this setting, the model is meta-learned over the entire 964 base classes defined in the original Omniglot dataset, and tested on the 659 classes in the test dataset, while incrementally exposing 10 classes per session starting with 10 classes and finishing with 600 classes.

\begin{figure*}[ht!]
  \centering \includegraphics[width=0.50\textwidth]{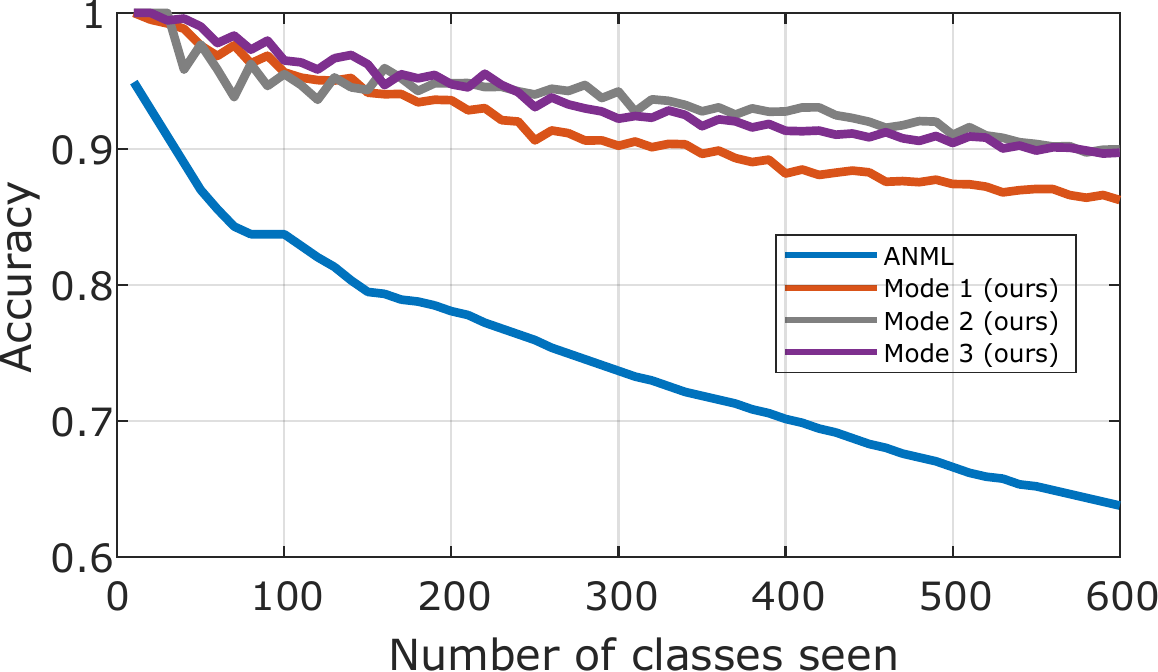}
  \captionsetup{width=.5\linewidth}
  \caption{Classification accuracy (\%) on Omniglot in the alternative FSCIL setting with $c$-way 5-shot where $c$ is the number of seen classes; ANML refers to the best performing model in~\cite{ecai2020}.}
  \label{fig:ecai_compare}
\end{figure*}

We compare our work with ANML~\cite{ecai2020} as the best performing model. 
%
The results are shown in Fig.~\ref{fig:ecai_compare}:
\name consistently performs better than ANML and minimizes the accuracy degradation, as more novel classes are incrementally added from 10 to 600.  
ANML incurs a drop of 31.1\% compared to 10.1\% in our Mode\,3.
This indicates the higher scalability of \name to cover a large number of classes in its lifespan.

\begin{table*}[h!]
\centering
\caption{Dimension ablation on miniImageNet. Classification accuracy (\%) of \name in the 5-way 5-shot FSCIL setting. }\label{tab:miniimagenet_d}
\begin{tabular}{lrccccccccc}
\toprule
\multicolumn{2}{l}{Session ($s$)}                         & 1     & 2     & 3     & 4     & 5     & 6     & 7     & 8    & 9                      \\
\multicolumn{2}{l}{No. of classes $|\tilde{\mathcal{C}}^{(s)}|$}   & 60     & 65     & 70    & 75     & 80     & 85     & 90     & 95    & 100                          \\
\cmidrule(r){1-2} \cmidrule(r){3-11} 
Mode & $d$ \\
\cmidrule(r){1-2} \cmidrule(r){3-11} 
\multirow{5}{*}{Mode 1}& 512 & 76.37 & 70.94 & 66.36 & 62.64 & 59.31 & 56.02 & 53.14 & 51.04 & 48.87                \\
                        & 256 & \textbf{76.78} & \textbf{71.08} & \textbf{66.37} & \textbf{62.75} & \textbf{59.33} & \textbf{56.39} & \textbf{53.34} & \textbf{51.11} & \textbf{48.94}     \\
                        & 128 & 76.32 & 70.72 & 65.93 & 62.16 & 58.63 & 55.74 & 52.83 & 50.73 & 48.48     \\
                        & 64  & 76.30 & 70.74 & 65.91 & 62.32 & 59.00 & 55.76 & 52.76 & 50.48 & 48.30     \\
                        & 32  & 74.10 & 68.58 & 63.86 & 60.23 & 56.95 & 53.67 & 50.80 & 48.44 & 46.33     \\
\cmidrule(r){1-2} \cmidrule(r){3-11}
\multirow{5}{*}{Mode 2}& 512 & 76.45 & \textbf{71.23} & \textbf{66.71} & \textbf{63.01} & \textbf{60.09} & 56.73 & 53.94 & 52.01 & 50.08 \\
                        & 256  & \textbf{76.70} & 70.95 & 66.19 & 62.80 & 59.65 & \textbf{56.80} & \textbf{54.29} & \textbf{52.08} & \textbf{50.58} \\
                        & 128  & 76.23 & 70.25 & 65.33 & 62.24 & 59.49 & 56.95 & 53.91 & 51.67 & 49.65 \\
                        & 64  & 75.83 & 68.54 & 62.57 & 58.59 & 56.09 & 53.49 & 50.62 & 48.66 & 46.95 \\
                        & 32  & 73.38 & 66.68 & 61.90 & 58.37 & 54.72 & 51.06 & 48.69 & 46.71 & 44.73 \\
\cmidrule(r){1-2} \cmidrule(r){3-11}
\multirow{5}{*}{Mode 3} & 512 & 76.40 & 71.14 & 66.46 & 63.29 & 60.42 & 57.46 & 54.78 & 53.11 & 51.41 \\
 & 256 & \textbf{76.75} & \textbf{71.17} & \textbf{66.50} & \textbf{63.39} & \textbf{60.86} & 58.05 & 55.30 & 53.08 & 51.41\\
 & 128 & 76.25 & 70.51 & 65.80 & 63.29 & 60.72 & \textbf{58.18} & \textbf{55.63} & \textbf{53.44} & \textbf{51.46}\\
 & 64  & 76.30 & 70.55 & 65.36 & 62.49 & 59.76 & 57.01 & 54.00 & 51.51 & 49.41\\
 & 32 & 73.97 & 67.82 & 62.61 & 59.20 & 56.25 & 52.52 & 49.54 & 47.41 & 45.99\\
\bottomrule
\end{tabular}
\end{table*}

\begin{table*}[h!]
\centering
\caption{Dimension ablation on CIFAR100. Classification accuracy (\%) of \name in the 5-way 5-shot FSCIL setting. }\label{tab:cifar100_d}
\begin{tabular}{lrccccccccc}
\toprule
\multicolumn{2}{l}{Session ($s$)}             & 1     & 2     & 3     & 4     & 5     & 6     & 7     & 8    & 9                      \\
\multicolumn{2}{l}{No. of classes $|\tilde{\mathcal{C}}^{(s)}|$}                      & 60     & 65     & 70    & 75     & 80     & 85     & 90     & 95    & 100                          \\
\cmidrule(r){1-2} \cmidrule(r){3-11} 
Mode & $d$ \\
\cmidrule(r){1-2} \cmidrule(r){3-11} 
\multirow{5}{*}{Mode 1}& 512     & \textbf{77.47} & \textbf{72.20} & \textbf{67.53} & \textbf{63.23} & \textbf{59.58} & \textbf{56.67} & \textbf{53.94} & \textbf{51.55} & \textbf{49.36}     \\
 & 256       &  77.10 & 72.07 & 67.48 & 63.22 & 59.56 & 56.52 & 53.87 & 51.31 & 49.10    \\
 & 128        &  76.95 & 71.78 & 67.09 & 62.75 & 59.08 & 56.35 & 53.63 & 51.24 & 48.72   \\
 & 64       &  76.92 & 71.50 & 66.72 & 62.46 & 58.74 & 55.65 & 52.86 & 50.38 & 48.23  \\
 & 32       &  74.85 & 69.82 & 64.91 & 60.33 & 56.99 & 53.92 & 51.32 & 48.87 & 46.75  \\
\cmidrule(r){1-2} \cmidrule(r){3-11}
 \multirow{5}{*}{Mode 2} & 512   & \textbf{77.50} & \textbf{72.45} &\textbf{67.94} & \textbf{63.80} & \textbf{60.24} & \textbf{57.34} & {54.61} & \textbf{52.41} & \textbf{50.23}  \\
 & 256   & 77.37 & 72.29 & 67.96 & 63.57 & 60.04 & 57.02 & \textbf{54.63} & 52.15 & 50.13   \\
 & 128   & 76.93 & 72.28 & 67.44 & 63.69 & 59.83 & 57.55 & 55.01 & 52.24 & 49.33   \\
 & 64   &  76.60 & 70.60 & 66.23 & 62.08 & 58.56 & 55.95 & 53.00 & 50.33 & 48.06  \\
 & 32   &  74.52 & 69.09 & 64.06 & 59.56 & 56.00 & 52.95 & 50.26 & 47.85 & 45.22  \\
\cmidrule(r){1-2} \cmidrule(r){3-11}
 \multirow{5}{*}{Mode 3} & 512    & \textbf{77.47} & \textbf{72.40} & 67.47 & 63.25 & 59.84 & 56.95 & 54.42 & 52.47 & 50.47 \\
 & 256    & 77.13 & 72.05 & \textbf{67.66} & \textbf{63.65} & \textbf{60.10} & 57.27 & 55.07 & 52.73 & 50.70   \\
 & 128    & 77.00 & 72.28 & 67.40 & 63.45 & 59.72 & \textbf{57.59} & \textbf{55.33} & \textbf{53.01} & \textbf{50.74}  \\
 & 64    & 77.00 & 71.45 & 67.26 & 63.00 & 59.75 & 56.94 & 54.41 & 51.92 & 49.20 \\
 & 32    & 74.92 & 68.98 & 64.20 & 58.81 & 55.41 & 52.56 & 50.04 & 47.41 & 45.33  \\
\bottomrule
\end{tabular}
\end{table*}

\begin{table*}[h!]
\centering
\caption{Dimension ablation on Omniglot. Classification accuracy (\%) on Omniglot in the 47-way 5-shot FSCIL setting. }\label{tab:omniglot_fscil_d}
\begin{tabular}{lrcccccccccc}
\toprule
\multicolumn{2}{l}{Session ($s$)}             & 1     & 2     & 3     & 4     & 5     & 6     & 7     & 8    & 9      & 10                \\
\multicolumn{2}{l}{No. of classes $|\tilde{\mathcal{C}}^{(s)}|$}                     & 1200 & 1247 & 1294 & 1341 & 1388 & 1435 & 1482 & 1529 & 1576 & 1623                     \\
\cmidrule(r){1-2} \cmidrule(r){3-12} 
Mode/ Work & $d$ \\
\cmidrule(r){1-2} \cmidrule(r){3-12} 
\multirow{5}{*}{Mode 1}& 512     & \textbf{84.16} & \textbf{83.82} & \textbf{83.69} & \textbf{83.32} & \textbf{83.22} & \textbf{82.78} & \textbf{82.70} & \textbf{82.32} & \textbf{81.77} & \textbf{81.56}    \\ 
 & 256          & 82.26 & 81.99 & 81.95 & 81.73 & 81.74 & 81.42 & 81.42 & 81.18 & 80.70 & 80.39  \\                  
 & 128          &  80.78 & 80.97 & 80.50 & 80.24 & 79.80 & 79.45 & 79.01 & 78.41 & 78.11 & 78.19\\
 & 64           & 72.89 & 72.87 & 72.49 & 72.18 & 71.52 & 70.85 & 70.73 & 69.87 & 69.59 & 69.30 \\          
 & 32           & 57.07 & 56.70 & 56.14 & 55.72 & 54.97 & 54.36 & 53.85 & 53.19 & 52.45 & 52.52 \\  
\cmidrule(r){1-2} \cmidrule(r){3-12}
 \multirow{5}{*}{Mode 2} & 512   & \textbf{86.87} & \textbf{86.77} & \textbf{86.57} & \textbf{86.44} & \textbf{86.40} & \textbf{86.20} & \textbf{86.25} & \textbf{85.96} & \textbf{85.63} & \textbf{85.49} \\
 & 256        & 84.32 & 84.35 & 84.23 & 83.94 & 84.02 & 83.93 & 83.86 & 83.78 & 83.44 & 83.19  \\
 & 128          &  82.85 & 83.29 & 82.70 & 82.65 & 82.14 & 82.03 & 81.91 & 81.31 & 80.68 & 81.15\\
 & 64           & 76.07 & 76.73 & 76.46 & 75.98 & 75.86 & 75.17 & 75.20 & 74.31 & 74.43 & 74.28 \\
 & 32           & 64.31 & 63.71 & 63.94 & 63.72 & 63.04 & 62.66 & 62.04 & 61.58 & 61.35 & 61.15 \\
\cmidrule(r){1-2} \cmidrule(r){3-12}
 \multirow{5}{*}{Mode 3} & 512   & \textbf{87.21} & \textbf{87.03} & \textbf{86.89} & \textbf{86.60} & \textbf{86.43} & \textbf{86.32} & \textbf{86.13} & \textbf{85.98} & \textbf{85.59} & \textbf{85.70} \\
 & 256          & 84.59 & 84.57 & 84.39 & 84.11 & 84.25 & 83.89 & 83.95 & 83.94 & 83.62 & 83.35 \\
 & 128          &  83.51 & 83.29 & 83.14 & 82.87 & 82.54 & 81.90 & 82.03 & 81.50 & 81.29 & 81.31\\
 & 64           & 76.67 & 76.30 & 76.46 & 76.22 & 75.66 & 75.51 & 75.53 & 74.19 & 74.23 & 74.25 \\
 & 32           & 64.99 & 64.66 & 64.61 & 63.72 & 63.34 & 63.14 & 62.66 & 62.23 & 61.69 & 62.17 \\
 \cmidrule(r){1-2} \cmidrule(r){3-12} 
\multirow{5}{*}{ProtoNet~\cite{snell17}} & 256 & \textbf{70.61} & \textbf{70.20} & \textbf{70.01} & \textbf{69.68} & \textbf{69.48} & \textbf{68.99} & \textbf{68.74} & \textbf{68.07} & \textbf{67.60} & -       \\
 & 128                  & 63.75 & 63.34 & 63.15 & 62.44 & 62.38 & 62.00 & 61.61 & 61.29 & 60.68 & 60.13   \\
 & 64                   & 49.69 & 49.10 & 48.63 & 48.13 & 47.67 & 46.97 & 46.73 & 46.11 & 45.47 & 45.08   \\      
 & 32                   & 36.53 & 36.11 & 35.74 & 35.45 & 34.87 & 34.37 & 34.08 & 33.42 & 32.99 & 32.71   \\
 \cmidrule(r){1-2} \cmidrule(r){3-12} 
 \multirow{5}{*}{CEC~\cite{Evolv_FSCIL_CVPR2021}}  & 512 & 76.44 & 76.62 & 76.21 & 76.10 & 75.37 & 74.92 & 74.60 & 74.04 & 73.43 & 73.19  \\
 & 256 & 76.94 & 76.94 & 76.56 & 76.35 & 75.62 & 75.20 & 74.84 & 74.45 & 73.94 & 73.59       \\
 & 128 & 77.10 & 77.15 & 76.94 & 76.89 & 76.26 & 75.79 & 75.46 & 75.05 & 74.58 & 74.28   \\
 & 64  & \textbf{78.91} & \textbf{79.07} & \textbf{78.74} & \textbf{78.60} & \textbf{77.94} & \textbf{77.55} & \textbf{77.18} & \textbf{76.77} & \textbf{76.39} & \textbf{76.11}  \\      
 & 32  & 74.51 & 74.59 & 74.32 & 73.97 & 73.31 & 72.76 & 72.28 & 71.84 & 71.55 & 71.41\\
\bottomrule
\end{tabular}
\end{table*}

\begin{table*}[h!]
\centering
\caption{Attention ablation on miniImageNet. Classification accuracy (\%) of \name with $d=512$ in the 5-way 5-shot FSCIL setting.}\label{tab:miniimagenet_attention}
\begin{tabular}{llccccccccc}
\toprule
\multicolumn{2}{l}{Session ($s$)}   & 1     & 2     & 3     & 4     & 5     & 6     & 7     & 8    & 9                      \\
\multicolumn{2}{l}{No. of classes $|\tilde{\mathcal{C}}^{(s)}|$}   & 60     & 65     & 70    & 75     & 80     & 85     & 90     & 95    & 100                          \\
Mode & Attention \\
\cmidrule(r){1-2} \cmidrule(r){3-11}
Mode\,1 & softabs                                     & 76.37 & \textbf{70.94} & \textbf{66.36} & \textbf{62.64} & \textbf{59.31} & \textbf{56.02} & \textbf{53.14} & \textbf{51.04} & \textbf{48.87}                \\
Mode\,1 & softmax                      &  \textbf{76.48} & \textbf{70.94} & \textbf{66.36} & 62.43 & 58.94 & 55.67 & 52.56 & 50.21 & 47.91               \\
\cmidrule(r){1-2} \cmidrule(r){3-11}
{Mode\,2} & softabs  & {76.45} & \textbf{71.23} & \textbf{66.71} & \textbf{63.01} & \textbf{60.09} & \textbf{56.73} & \textbf{53.94} & \textbf{52.01} & \textbf{50.08} \\
{Mode\,2} & softmax   & \textbf{76.48} & 71.18 & 66.67 & 62.80 & 59.54 & 56.33 & 53.04 & 51.04 & 49.09 \\
\cmidrule(r){1-2} \cmidrule(r){3-11}
{Mode\,3} & softabs                       & 76.40 & \textbf{71.14} &\textbf{66.46} & \textbf{63.29} & \textbf{60.42 }& \textbf{57.46} & \textbf{54.78} & \textbf{53.11} & \textbf{51.41} \\
{Mode\,3} & softmax                       & \textbf{76.47} & 70.86 & 65.90 & 62.53 & 59.72 & 56.88 & 54.47 & 51.83 & 50.05 \\
\bottomrule
\end{tabular}
\end{table*}

\begin{table*}[h!]
\centering
\caption{Attention ablation on CIFAR100. Classification accuracy (\%) of \name with $d=512$ in the 5-way 5-shot FSCIL setting.}\label{tab:cifar100_attention}
\begin{tabular}{llccccccccc}
\toprule
\multicolumn{2}{l}{Session ($s$)}   & 1     & 2     & 3     & 4     & 5     & 6     & 7     & 8    & 9                      \\
\multicolumn{2}{l}{No. of classes $|\tilde{\mathcal{C}}^{(s)}|$}   & 60     & 65     & 70    & 75     & 80     & 85     & 90     & 95    & 100                          \\
Mode & Attention \\
\cmidrule(r){1-2} \cmidrule(r){3-11}
Mode\,1 & softabs     & \textbf{77.47} & \textbf{72.20} & \textbf{67.53} & \textbf{63.23} & \textbf{59.58} & \textbf{56.67} & \textbf{53.94} & \textbf{51.55} & \textbf{49.36}     \\
Mode\,1 & softmax    &  76.35 & 71.03 & 66.31 & 62.13 & 58.45 & 55.40 & 52.69 & 50.20 & 47.99    \\
\cmidrule(r){1-2} \cmidrule(r){3-11}
Mode\,2 & softabs    & \textbf{77.50} & \textbf{72.45} & \textbf{67.94} & \textbf{63.80} & \textbf{60.24} & \textbf{57.34} & \textbf{54.61} & \textbf{52.41} & \textbf{50.23}   \\
Mode\,2 & softmax  & 76.33 & 71.09 & 66.49 & 62.16 & 58.8 & 55.78 & 53.17 & 50.75 & 48.39\\
\cmidrule(r){1-2} \cmidrule(r){3-11}
Mode\,3 & softabs   & \textbf{77.47} & \textbf{72.40} & \textbf{67.47} & \textbf{63.25} & \textbf{59.84} & \textbf{56.95} & \textbf{54.42} & \textbf{52.47} & \textbf{50.47} \\
Mode\,3 & softmax   & 76.35 & 70.88 & 65.96 & 61.99 & 58.17 & 55.00 & 52.38 & 49.92 & 47.81 \\
\bottomrule
\end{tabular}
\end{table*}

\begin{table*}[h!]
\centering
\caption{
Attention ablation on Omniglot. Classification accuracy (\%) of \name with $d=512$ in the 47-way 5-shot FSCIL setting.
}\label{tab:omniglot_attention}
\begin{tabular}{llcccccccccc}
\toprule
\multicolumn{2}{l}{Session ($s$)}   & 1     & 2     & 3     & 4     & 5     & 6     & 7     & 8    & 9    & 10                  \\
\multicolumn{2}{l}{No. of classes $|\tilde{\mathcal{C}}^{(s)}|$}   & 1200 & 1247 & 1294 & 1341 & 1388 & 1435 & 1482 & 1529 & 1576 & 1623                     \\
Mode & Attention \\
\cmidrule(r){1-2} \cmidrule(r){3-12}
Mode\,1 & softabs    &      84.16 & 83.82 & 83.69 & 83.32 & 83.22 & 82.78 & 82.70 & 82.32 & 81.77 & 81.56    \\ 
Mode\,1 & softmax    &      \textbf{85.42} & \textbf{85.15} & \textbf{85.05} & \textbf{84.69} & \textbf{84.57} & \textbf{84.22} & \textbf{84.11} & \textbf{83.94} & \textbf{83.57} & \textbf{83.54}  \\
\cmidrule(r){1-2} \cmidrule(r){3-12}
Mode\,2 & softabs    &      86.87 & \textbf{86.77} & 86.57 & \textbf{86.44} & \textbf{86.40} & \textbf{86.20} & \textbf{86.25} & \textbf{85.96} & \textbf{85.63} & \textbf{85.49} \\
Mode\,2 & softmax    &      \textbf{86.93} & 86.69 & \textbf{86.60} & 86.42 & 86.17 & 86.00 & 86.03 & 85.78 & 85.40 & 85.36  \\
\cmidrule(r){1-2} \cmidrule(r){3-12}
Mode\,3 & softabs    &      \textbf{87.21} & \textbf{87.03} & \textbf{86.89} & \textbf{86.60} & \textbf{86.43} & \textbf{86.32} & \textbf{86.13} & \textbf{85.98} & \textbf{85.59} & \textbf{85.70} \\
Mode\,3 & softmax    &      87.01 & 86.89 & 86.80 & 86.57 & 86.29 & 86.08 & 86.02 & 85.78 & 85.36 & 85.33 \\
\bottomrule
\end{tabular}
\end{table*}

\begin{table*}[h!]
\centering
\caption{Feature extractor ablation on miniImageNet. Classification accuracy (\%) in the 5-way 5-shot FSCIL setting.}\label{tab:miniimagenet_FE}
\begin{tabular}{llccccccccc}
\toprule
\multicolumn{2}{l}{Session ($s$)}   & 1     & 2     & 3     & 4     & 5     & 6     & 7     & 8    & 9                      \\
\multicolumn{2}{l}{No. of classes $|\tilde{\mathcal{C}}^{(s)}|$}   & 60     & 65     & 70    & 75     & 80     & 85     & 90     & 95    & 100                          \\
Mode/ Work & Feature extractor \\
\cmidrule(r){1-2} \cmidrule(r){3-11}
AL-MML~\cite{FSCIL_CVPR2020}  & ResNet-18  & 61.31 & 50.09 & 45.17 & 41.16 & 37.48 & 35.52 & 32.19 & 29.46 & 24.42             \\
IDLVQ-C~\cite{VecQ_FSCIL_ICLR2021} & ResNet-18 & 64.77 & 59.87 & 55.93 & 52.62 & 49.88 & 47.55 & 44.83 & 43.14 & 41.84\\
Semantic KD~\cite{Semantic_FSCIL_CVPR2021} & ResNet-18 & $<$62 & $<$59 & $<$54 & $<$50 & $<$49 & $<$45 & $<$42 & $<$40 & $<$39\\
VAE~\cite{Synthesized_FSCIL_ICCV2021} & ResNet-18 & $<$62 & $<$60 & $<$54 & $<$52 & $<$50 & $<$49 & $<$46 & $<$44 & $<$43\\  
F2M~\cite{FlatMin_FSCIL_NIPS2021} & ResNet-18 & 67.28 & 63.80 & 60.38 & 57.06 & 54.08 & 51.39 & 48.82 & 46.58 & 44.65\\ 
CEC~\cite{Evolv_FSCIL_CVPR2021} & ResNet-18  & 72.00 & 66.83 & 62.97 & 59.43 & 56.70 & 53.73 & 51.19 & 49.24 & 47.63               \\
\cmidrule(r){1-2} \cmidrule(r){3-11}
{\name Mode\,1}  & ResNet-12                                   & 76.37 & 70.94 & 66.36 & 62.64 & 59.31 & 56.02 & 53.14 & 51.04 & 48.87                \\
{\name Mode\,2}  & ResNet-12                      & {76.45} & {71.23} & {66.71} & 63.01 & 60.09 & 56.73 & 53.94 & 52.01 & 50.08 \\
{\name Mode\,3}  & ResNet-12                      & 76.40 & 71.14 & 66.46 & {63.29} & {60.42 }& {57.46} & {54.78} & {53.11} & {51.41} \\
\cmidrule(r){1-2} \cmidrule(r){3-11}
\name Mode 1 & ResNet-12 (small) & 76.08 & 70.63 & 66.11 & 62.23 & 58.91 & 56.12 & 53.11 & 51.02 & 48.93             \\
\name Mode 2 & ResNet-12 (small) & 75.90 & 70.52 & 66.01 & 62.11 & 58.86 & 56.19 & 53.23 & 51.31 & 49.53\\
\name Mode 3 & ResNet-12 (small) & 76.12 & 70.20 & 65.29 & 62.25 & 59.35 & 56.76 & 54.18 & 52.15 & 50.47              \\
\bottomrule
\end{tabular}
\end{table*}

\begin{table*}[h!]
\centering
\caption{Feature extractor ablation on CIFAR100. Classification accuracy (\%) in the 5-way 5-shot FSCIL setting.}\label{tab:cifar100_FE}
\begin{tabular}{llccccccccc}
\toprule
\multicolumn{2}{l}{Session ($s$)}   & 1     & 2     & 3     & 4     & 5     & 6     & 7     & 8    & 9                      \\
\multicolumn{2}{l}{No. of classes $|\tilde{\mathcal{C}}^{(s)}|$}   & 60     & 65     & 70    & 75     & 80     & 85     & 90     & 95    & 100                          \\
Mode/ Work & Feature extractor \\
\cmidrule(r){1-2} \cmidrule(r){3-11}
AL-MML~\cite{FSCIL_CVPR2020}  & ResNet-18  & 64.10 & 55.88 & 47.07 & 45.16 & 40.11 & 36.38 & 33.96 & 31.55 & 29.37           \\
Semantic KD$^{*}$~\cite{Semantic_FSCIL_CVPR2021} & ResNet-18& $<$64 & $<$57 & $<$51 & $<$46 & $<$43 & $<$41 & $<$39 & $<$37 & $<$35   \\
VAE$^{*}$~\cite{Synthesized_FSCIL_ICCV2021} & ResNet-18& $<$62 & $<$58 & $<$57 &$<$52 &$<$51 & $<$49 & $<$46 & $<$45 & $<$42\\ 
F2M~\cite{FlatMin_FSCIL_NIPS2021} & ResNet-18& 64.71 & 62.05 & 59.01 & 55.58 & 52.55 & 49.96 & 48.08 & 46.67 & 44.67 \\ 
CEC~\cite{Evolv_FSCIL_CVPR2021}   & ResNet-20 & 73.07 & 68.88 & 65.26 & 61.19 & 58.09 & 55.57 & 53.22 & 51.34 & 49.14     \\
\cmidrule(r){1-2} \cmidrule(r){3-11}
{\name Mode\,1}      & ResNet-12     & 77.47 & 72.20 & 67.53 & 63.23 & 59.58 & 56.67 & 53.94 & 51.55 & 49.36     \\
{\name Mode\,2 }     & ResNet-12                                 & {77.50} & {72.45} & {67.94} & {63.80} & {60.24} & {57.34} & {54.61} & 52.41 & 50.23   \\
{\name Mode\,3 }  & ResNet-12   & 77.47 & 72.40 & 67.47 & 63.25 & 59.84 & 56.95 & 54.42 & {52.47} & {50.47} \\
\cmidrule(r){1-2} \cmidrule(r){3-11}
\name Mode\,1 & ResNet-12 (small) &     76.58 & 71.51 & 66.79 & 62.49 & 58.8 & 55.72 & 52.91 & 50.56 & 48.39 \\
\name Mode\,2 & ResNet-12 (small)  &  76.57 & 71.86 & 67.34 & 63.05 & 59.46 & 56.42 & 53.80 & 51.37 & 49.26  \\
\name Mode\,3 & ResNet-12 (small)  & 76.58 & 71.74 & 66.71 & 62.20 & 58.94 & 56.21 & 53.63 & 51.41 & 49.50\\
\bottomrule
\end{tabular}
\end{table*}

\begin{table*}[h!]
\centering
\caption{Memory compression on miniImageNet. Classification accuracy (\%) of \name in the 5-way 5-shot FSCIL setting.}\label{tab:miniimagenet_compression}
\begin{tabular}{llccccccccc}
\toprule
\multicolumn{2}{l}{Session ($s$)}   & 1     & 2     & 3     & 4     & 5     & 6     & 7     & 8    & 9                      \\
\multicolumn{2}{l}{No. of classes $|\tilde{\mathcal{C}}^{(s)}|$}   & 60     & 65     & 70    & 75     & 80     & 85     & 90     & 95    & 100                          \\
Mode & Compression \\
\cmidrule(r){1-2} \cmidrule(r){3-11}
Mode\,1 & No compression                                     & 76.37 & {70.94} & {66.36} & {62.64} & {59.31} & {56.02} & {53.14} & {51.04} & {48.87}                \\
Mode\,1 & 2$\times$ compressing EM &  74.65 & 69.31 & 64.10 & 60.43 & 56.84 & 53.51 & 49.94 & 48.05 & 45.34  \\ 
\cmidrule(r){1-2} \cmidrule(r){3-11}
{Mode\,3} & No compression                       & 76.40 & {71.14} &{66.46} & {63.29} & {60.42 }& {57.46} & {54.78} & {53.11} & {51.41} \\
{Mode\,3} & 2$\times$ compressing GAAM  & 71.72 & 66.40 & 61.41 & 57.13 & 53.56 & 50.38 & 47.74 & 45.28 & 42.91 \\ 
\bottomrule
\end{tabular}
\end{table*}

\end{document}